\documentclass[10pt,twocolumn,letterpaper]{article}
\usepackage[accsupp]{axessibility}
\usepackage{iccv}
\usepackage{times}
\usepackage{epsfig}
\usepackage{graphicx}
\usepackage{amsmath}
\usepackage{amssymb}

\usepackage{booktabs}
\usepackage{pifont}
\usepackage{caption}
\usepackage{array}
\usepackage{color}
\usepackage[x11names]{xcolor}
\usepackage{multirow}
\usepackage{diagbox}
\usepackage{booktabs}
\usepackage{siunitx}
\usepackage{appendix}
\usepackage{etoc}
\usepackage{subcaption}

\usepackage[pagebackref=true,breaklinks=true,letterpaper=true,colorlinks,bookmarks=false]{hyperref}

\newcommand{\gt}[0]{GT\xspace}
\newcommand{\rgb}[0]{RGB\xspace}
\newcommand{\rgbd}[0]{RGB-D\xspace}
\newcommand{\Ours}[0]{SHOWMe\xspace}
\newcommand{\ho}[0]{HO\xspace}

\newcommand{\mvr}[0]{MVR\xspace}
\newcommand{\briac}[0]{FDR\xspace}
\newcommand{\sigg}[0]{HHOR\xspace}
\newcommand{\vh}[0]{VH\xspace}

\newcommand{\greencheck}{{\color{Green4} \checkmark}}
\newcommand{\redcross}{{\color{red}$\times$}}

\DeclareMathOperator{\sg}{sg}

\usepackage[capitalize]{cleveref}
\crefname{section}{Sec.}{Secs.}
\Crefname{table}{Table}{Tables}
\crefname{table}{Tab.}{Tabs.}   

\iccvfinalcopy 

\def\iccvPaperID{39} 

\ificcvfinal\fi

\begin{document}

\title{SHOWMe: Benchmarking Object-agnostic Hand-Object 3D Reconstruction}
\author{Anilkumar Swamy$^{1,2}$ \hspace{0.1cm} Vincent Leroy$^{1}$ \hspace{0.1cm} Philippe Weinzaepfel$^{1}$ \hspace{0.1cm}  Fabien Baradel$^{1}$ \hspace{0.1cm} Salma Galaaoui$^{1}$ 
\and
\hspace{0.1cm} Romain Brégier$^{1}$ \hspace{0.1cm} Matthieu Armando$^{1}$ \hspace{0.1cm} Jean-Sebastien Franco$^{2}$   \hspace{0.1cm} Grégory Rogez$^{1}$
\and
\textsuperscript{1}\footnotetext[1]{}{}{NAVER LABS Europe} 
\textsuperscript{2}\footnotetext[1]{}{}{Inria centre at the University Grenoble Alpes}\\
}
\ificcvfinal\thispagestyle{empty}\fi

\twocolumn[{%
\renewcommand\twocolumn[1][]{#1}%
\maketitle
\begin{center}
    \captionsetup{type=figure}
    \includegraphics[width=\linewidth]{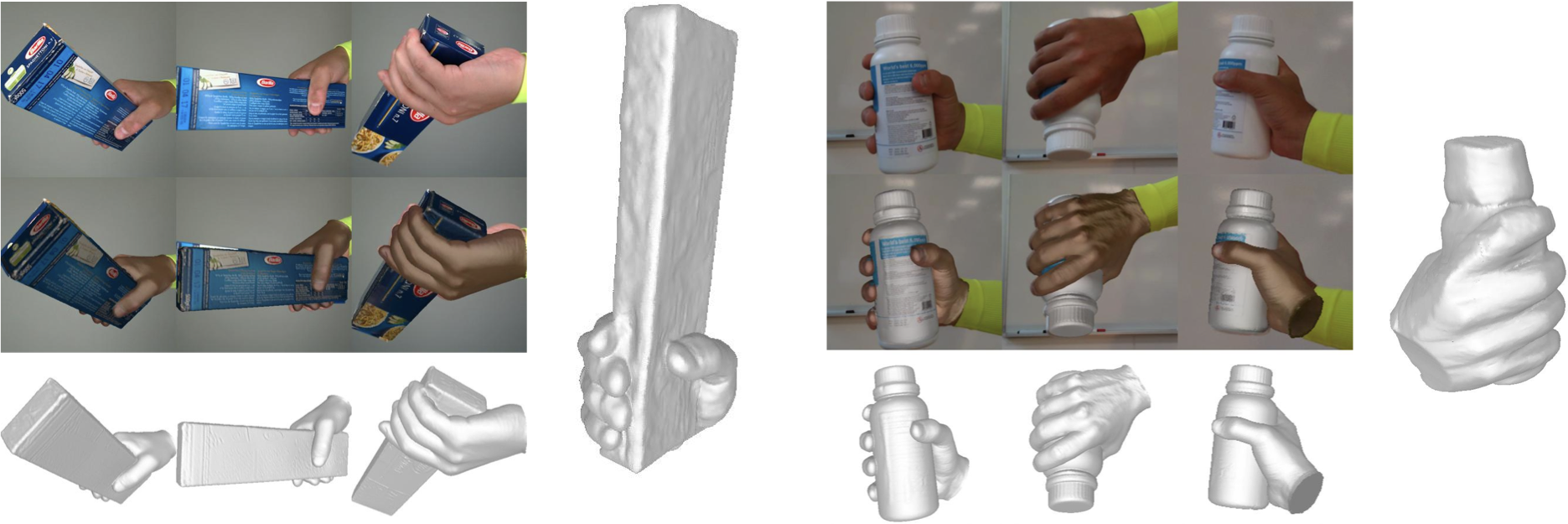} \\
    [-0.38cm]
    \captionof{figure}{ \textbf{The  \Ours dataset} comprises 96 videos with their associated high-quality textured meshes of a hand holding an object.
    For two different samples,  we show on the left side, row by row,  real RGB crops from the dataset, an overlay of the corresponding ground truth textured mesh,    and a rendering of the texture-less mesh with Phong shading. On the right, we show the 3D reconstruction of the hand-object obtained from the \rgb stream only, using one of the evaluated baselines. 
    }
    \label{fig:teaser}
\end{center}
}]

\begin{abstract}

Recent hand-object interaction datasets show limited real object variability and rely on fitting the MANO parametric model to obtain groundtruth hand shapes. To go beyond these limitations and spur further research, we introduce the \Ours dataset which consists of 96 videos, annotated with real and detailed hand-object 3D textured meshes.
Following recent work, we consider a rigid hand-object scenario, in which the pose of the hand with respect to the object remains constant during the whole video sequence. This assumption allows us to register sub-millimetre-precise groundtruth 3D scans to the image sequences in \Ours.  Although simpler, this hypothesis makes sense in terms of applications where the required accuracy and level of detail is important \eg, object hand-over in human-robot collaboration, object scanning, or manipulation and contact point analysis.  Importantly, the rigidity of the hand-object systems allows to tackle video-based 3D reconstruction of unknown hand-held objects using a 2-stage pipeline consisting of a rigid registration step followed by a multi-view reconstruction (MVR) part. We carefully evaluate a set of non-trivial baselines for these two stages and show that it is possible to achieve promising object-agnostic 3D hand-object reconstructions employing an SfM toolbox or a hand pose estimator to recover the rigid transforms and off-the-shelf MVR algorithms.  However, these methods remain sensitive to the initial camera pose estimates which might be imprecise due to lack of textures on the objects or heavy occlusions of the hands, leaving room for improvements in the reconstruction. Code and dataset are available at \href{https://europe.naverlabs.com/research/showme/}{https://europe.naverlabs.com/research/showme/}.
\end{abstract}

 \begin{figure*}[ht]
 \centering
 \includegraphics[width=\linewidth]{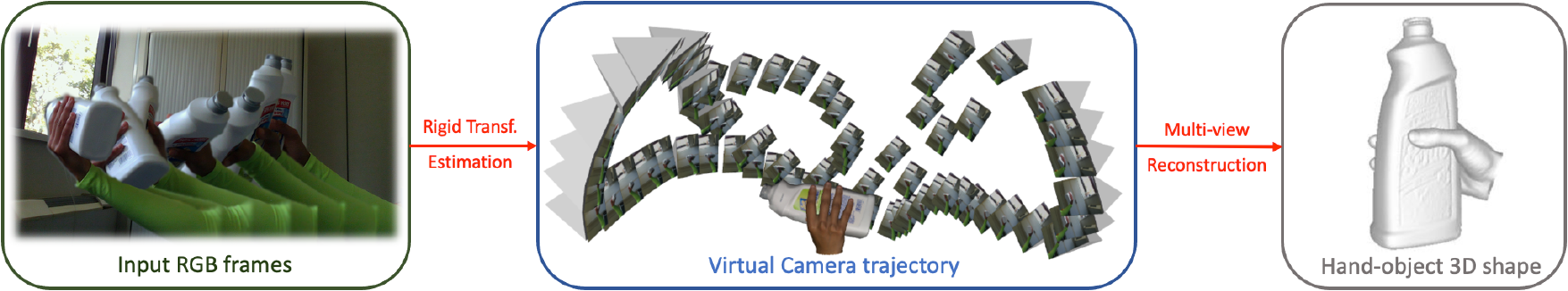} \\[-0.31cm]
 \caption{ \textbf{Hand-Object 2-stage 3D reconstruction  pipeline.} Given an RGB video of a hand holding an object (left), 
 the rigid transformation between frames is first estimated.
 This allows to see the problem as if 
 a set of multiple virtual cameras observe
 a fixed hand-object system (middle). Multi-view reconstruction can then be employed to estimate an accurate hand-object 3D shape (right).  
We benchmark several baselines for both stages using the presented dataset.
 }

\vspace{-0.35cm}
 
 \label{fig:overviewTask}
 \end{figure*}
 
\section{Introduction}
\label{sec:intro}

Understanding interactions between  hands and objects from \rgb images is a key component towards  better understanding  human actions and interactions. 
Such understanding could benefit many applications,
from virtual and augmented reality to human-robot
interaction and autonomous robotic manipulation via learning by demonstration. For instance, in a scenario where a human is  handing over an object to a robot equipped with \rgb sensors, 
we expect the robot to grasp the object without hurting the person 
in any way.
Such action is likely to require a fine-grained perception of both the object and the hand holding it, and being able to accurately model the hand-object~(\ho) system in 3D from \rgb data would be very useful in such context.

This problem of joint \ho 3D reconstruction has been addressed in a large body of recent works  ~\cite{hasson_learning_2019,hasson_leveraging_2020,hasson_towards_2022,contactpose,hampali2020honnotate,cao_reconstructing_2021,chen_alignsdf_2022, ye_whats_2022,chen_alignsdf_2022,tse2022collaborative} that estimate \ho 3D shape from single \rgb images. These methods often rely on  a deformable kinematic model of the human hand, MANO~\cite{romero_embodied_2017},  
which contains useful priors, but also limits the potential reconstruction accuracy~\cite{lisa} for unseen hand shapes.
A second important limitation of most HO reconstruction approaches is that the exact 3D model of the object is often assumed to be known apriori, and they tend to struggle to generalize to objects that fall outside of the training distribution. 
While single-image \ho reconstruction without priors over the objects remains very challenging, 
exploiting multiple observations of the scene can significantly simplify the task.

One way to obtain more observations is to consider a synchronized multi-camera setup, increasing the complexity of  deployment in practice.  
Another way is to focus on the temporal aspect of the \rgb streams as in \cite{huang2022hhor} who recently showed that multiple observations of the scene can be exploited to simplify object-agnostic hand-object 3D reconstruction. However, their method remains limited to close-up fingertip grasps of small objects and cannot be used for natural hand-object interactions. 
Interestingly, seldom previous work focused on aggregating temporal information of a \rgb  video for \ho reconstruction~\cite{huang2022hhor}, unless the strong assumption of a known object was made~\cite{hasson_towards_2022,hasson_leveraging_2020}.

Following \cite{contactpose,huang2022hhor}, we simplify the problem as an intermediary step towards dynamic temporal integration by assuming
that the camera is static and the hand is holding an unknown object rigidly.
In this setup, an \rgb video can be viewed as multiple observations of the same \ho system, which allows to formulate the \ho modeling problem in a Multi-View Reconstruction (\mvr) setting: the \rgb appearance of a \ho instance that undergoes a rigid transformations is observed. In order to solve this problem, two unknowns have to be addressed: 1. the rigid transformation and 2. how to aggregate \rgb observations. It is worth noting that these points can be addressed either separately or jointly.  With the exception of \cite{huang2022hhor} who operates in a rather constrained scenario, no method was specifically designed to solve the challenges raised by this task but, more importantly, there is a need for an evaluation protocol and a specifically designed dataset.

Therefore, we propose a novel dataset consisting of 96 videos of a hand holding an object rigidly and showing this object to the camera. We captured a total of 87K frames depicting 42 unique objects with evenly distributed grasp configurations, handled by 15
subjects reflecting a diversity of gender, color, and hand shape.  
Importantly, our dataset contains high-precision ground-truth (GT) \ho 3D shapes, that we captured using a sub-millimeter precision scanner before capturing each video sequence. The resulting textured 3D meshes are then registered to each frame of the corresponding sequences, in order to provide highly detailed GT annotations.
In practice, we proceed in two steps:
1) we register the GT \ho mesh to the depth map of each frame in the sequence. 
2) We refine the registration using a differentiable rendering pipeline to obtain very accurate alignments of the 3D mesh with the \rgb frames as shown in \cref{fig:teaser}.
We call our dataset \Ours, standing for Single-camera Hand-Object videos With accurate textured 3D Meshes. 
\begin{table*}[ht]
\centering
\resizebox{1.00\linewidth}{!}{
\begin{tabular}{lcccccccccccc}
\toprule
\multirow{2}{*}{dataset} & real  & marker- & \multicolumn{4}{c}{\# number of}  & image   & grasp   & object & hand-obj  & hand & hand       \\ \cline{4-7}
 &
  \multicolumn{1}{c}{images} &
  \multicolumn{1}{c}{less} &
  \multicolumn{1}{c}{img} &
  \multicolumn{1}{c}{seq} &
  \multicolumn{1}{c}{sbj} &
  \multicolumn{1}{c}{obj} &
  \multicolumn{1}{c}{resol.} &
  \multicolumn{1}{c}{variability} &
  \multicolumn{1}{c}{scan} &
  \multicolumn{1}{c}{texture} &
  \multicolumn{1}{c}{scan} &
  \multicolumn{1}{c}{annotation} \\ \hline 
  ObMan\cite{hasson_learning_2019}  & \redcross & \greencheck & 154k & - & 20 & 3K &   256 $\times$ 256 & +++ & \greencheck & \redcross & \redcross & MANO \\
  GRAB\cite{grab} & \redcross & \redcross & - & 1,335 & 10 & 51 &  - &  +++ & \greencheck & \redcross & \redcross & MANO \\
  \hline
  FPHA\cite{garcia-hernando_first-person_2018}  & \greencheck  & \redcross  & 105k  & 1,175                      & 6 & 4   &  1920$\times$1080   & + & \greencheck      & \redcross & \redcross   & keypoints   \\
  ContactPose\cite{contactpose}  & \greencheck & \redcross & 2,991k & 2,303 & 50 & 25  & 960$\times$540 & ++ & \greencheck & \redcross & \redcross & MANO \\ 
  
  ARCTIC\cite{fan_articulated_2022}  & \greencheck & \redcross & 1,200k & 242 & 9 & 10 &  2800 $\times$ 2000 & +++ & \greencheck & \redcross & \redcross & MANO\\
  \hline
  YCB-Affordance\cite{Ganhand}  & \greencheck & \greencheck & 133k & - & 1 & 21 & 640 $\times$ 480 & +++ & \greencheck &  \redcross & \redcross & MANO \\
  GUN-71\cite{rogez_understanding_2015}  & \greencheck  & \greencheck  & 12k  & 1,680 & 8 & 1988  & 640$\times$480   & +++ & \redcross & \redcross & \redcross & grasp Id \\
  FreiHand\cite{zimmermann_freihand_2019}  & \greencheck  & \greencheck & 37k  & -                      & 32 & 27    & 224$\times$224   & ++ & \redcross      & \redcross & \redcross   & MANO    \\ 

  Dexter+Object\cite{sridhar_real-time_2016} & \greencheck  & \greencheck & 3k  & 6                      & 2 & 2    & 640$\times$480   & + & \redcross      & \redcross & \redcross   & fingertips    \\ 

  EgoDexter\cite{mueller_real-time_nodate}  & \greencheck  & \greencheck & 3k  & 4                      & 4 & -    & 640$\times$480   & + & \redcross      & \redcross & \redcross   & fingertips    \\

HO3D\cite{hampali2020honnotate} & \greencheck & \greencheck & 78k & 27 & 10 & 10 & 640 $\times$ 480 & +++ &  \greencheck &  \redcross & \redcross & MANO \\

DexYCB\cite{chao_dexycb_2021}  & \greencheck & \greencheck & 582k & 1,000 & 10 & 20  & 640 $\times$ 480 & ++ & \greencheck &  \redcross & \redcross & MANO  \\

H2O\cite{kwon_h2o_2021} & \greencheck & \greencheck & 571k & - & 4 & 8  & 1280 $\times$ 720 & ++ & \greencheck &  \redcross & \redcross & MANO \\

OakInk\cite{yang_oakink_2022}  & \greencheck & \greencheck & 230k & - & 12 & 100  & 848 $\times$ 480 & +++ & \redcross & \redcross & \redcross  & MANO \\
\hline

HOD\cite{huang2022hhor} & \greencheck & \greencheck & 126k & 70 & \textcolor{red}{1} & 35  & 2160 $\times$ 3840 & \textcolor{red}{+} &  \greencheck \textcolor{red}{(only 14)} & \redcross & \redcross & \textcolor{red}{NO Annotations}  \\

\textbf{\Ours} (Ours)  & \greencheck & \greencheck & \textbf{87k} & \textbf{96} & \textbf{15} & \textbf{42} & \textbf{1280}$\times$\textbf{720} & +++ & \greencheck  & \greencheck & \greencheck & MANO \\
\bottomrule
\end{tabular}
}
\vspace{-0.3cm}
\caption{\textbf{Comparison of our dataset with existing hand-object interaction datasets}
}
\label{tab:compare_datasets}

\vspace{-0.3cm}

\end{table*}

Using \Ours, we  benchmark the
2-stage pipeline consisting of a rigid registration followed by a \ho  3D reconstruction from multiple observations, see~\cref{fig:overviewTask}. In the same spirit as \cite{ma2022virtual} with body shapes, 
we first estimate the rigid transformations between frames using the output of a hand keypoints detector as in~\cite{huang2022hhor}.
We compare this approach to a standard structure-from-motion (SfM) approach, namely  COLMAP~\cite{schoenberger2016sfm}. 
We find that hand-based estimation of the rigid transformation is more robust for textureless objects but suffers in case of heavy occlusions.
Given the rigid registration,  
the \ho reconstruction can be performed using multi-view reconstruction methods.
We compare a silhouette-based reconstruction method, leveraging hand-object segmentation~\cite{leroyvh} to more recent approaches based on differentiable rendering method~\cite{toussaint2022fast} and neural implicit surfaces~\cite{huang2022hhor}.
All three obtain extremely accurate results given ground-truth registration.
Yet, when considering estimated registrations, results of the best baseline are satisfactory on approximately three-quarters of the sequences and fail on the others. 
This confirms that HO 3D reconstruction from an RGB video is a difficult task, and we hope our dataset will foster further research on this topic.

In summary, our contribution is twofold. First, we propose a novel hand-object interaction dataset, \Ours, and the pipeline we designed to annotate \rgbd videos using high-precision \ho 3D scans. \Ours is the first dataset providing such level of accuracy for the GT hand-object 3D shapes. Second, we evaluate a set of baselines for the MVR-based pipeline for detailed and object-agnostic \ho 3D reconstruction in RGB videos.

After discussing related work and existing datasets in \cref{sec:related}, we introduce the \Ours dataset and its capturing setup in \cref{sec:dconstruct}. We finally present the 2-stage pipeline in \cref{sec:pipeline} before evaluating several baselines in \cref{sec:expe}.

\section{Related Work}
\label{sec:related}

Our two contributions being a new \ho  dataset  and a benchmark of object-agnostic \ho reconstruction baselines, we discuss below the most relevant datasets  
and  methods. 

\noindent \textbf{Hand-Object Datasets.}
Earlier research on hand-object interaction~\cite{rogez_understanding_2015, bambach_lending_2015, bullock_yale_2015, minjie_cai_scalable_2015, fathi_learning_2011} have proposed datasets for grasp classification or action recognition. Despite the importance of the recognition tasks, these datasets were seldom considered for \ho  reconstruction  research due to the unavailability of \gt 3D annotations, such as 3D joints or   3D shapes.

Obtaining images with ground-truth 3D information is a tedious problem in general, even for non-hand-related research. The small size of the hands in images make them difficult to annotate manually~\cite{sridhar_real-time_2016}. The problem is exacerbated when considering a hand interacting with objects. Past work has therefore proposed to consider synthetic data~\cite{RogezSR15, mueller_real-time_nodate,choi_robust_2017,hasson_learning_2019,Ganhand}, motion capture with markers~\cite{contactpose}, magnetic sensors~\cite{garcia-hernando_first-person_2018} or multi-view set-ups~\cite{zimmermann_freihand_2019, hampali2020honnotate, contactpose, chao_dexycb_2021,kwon_h2o_2021, yang_oakink_2022}. Synthetic data is usually obtained by rendering a parametric model of the hand interacting with objects. Even if realism is sufficient when considering a depth sensor~\cite{RogezSR15}, the domain gap between synthetic and real RGB images is often too large to be a valid option on its own. On the other hand, invasive motion capture methods based on magnetic sensors and markers make the hand appearance unrealistic and introduces an undesired bias.

Most of the recent datasets obtained through multi-view set-ups~\cite{zimmermann_freihand_2019, hampali2020honnotate, contactpose, chao_dexycb_2021,kwon_h2o_2021, yang_oakink_2022} use the multi-view data to fit the MANO parametric model~\cite{romero_embodied_2017} that is then considered as GT hand shape. Although it contains useful priors, MANO cannot represent very detailed hand shapes~\cite{lisa}. In our case, we scan the hand using a high-precision  scanner, obtaining a GT shape with sub-millimeter accuracy.

Recent multiview video datasets such as ~\cite{hampali2020honnotate, chao_dexycb_2021,hampali2021ho}  are impressive in terms of scale, markerless nature, and realism in motion but they lack object variability (10 objects for ~\cite{hampali2020honnotate} and 20 for~\cite{chao_dexycb_2021}, both object sets from the YCB dataset~\cite{calli_benchmarking_2015}). Motions are also limited to the same patterns like lifting the objects from the table and placing it back or handing them over to another person.  OakInk~ \cite{yang_oakink_2022} provides a much larger variety of objects but with limited motions. The \Ours dataset contains more than 40 objects with complex movements showing all sides of the object.
\begin{figure*}
\center
\includegraphics[width=\linewidth]{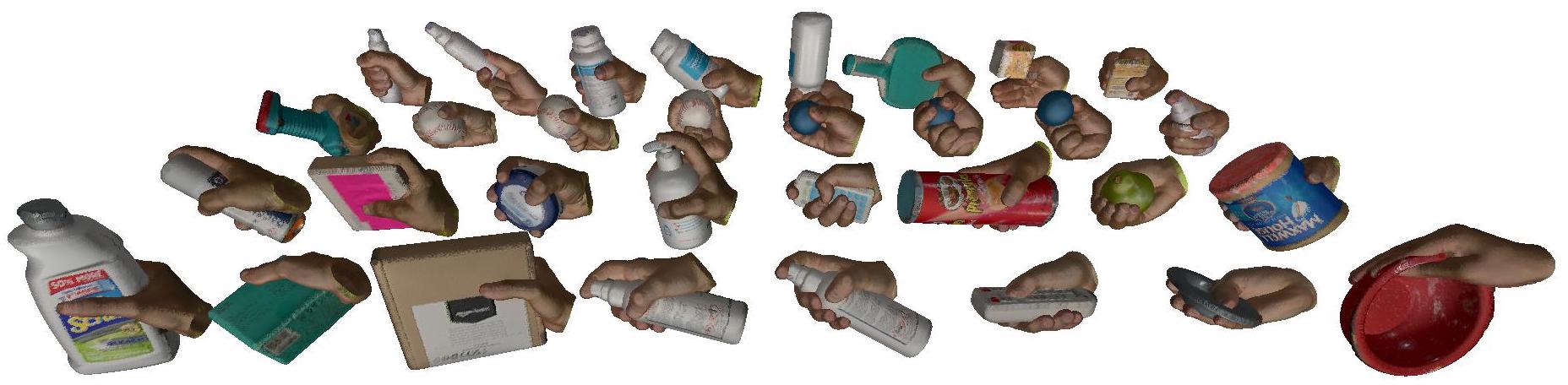} \\[-0.25cm]
\caption{
\textbf{Rendering of the textured mesh for few hand-object configurations of our \Ours dataset}.
}
\label{fig:mosaic_orbit}

\vspace{-0.25cm}

\end{figure*}

Closer to the proposed \Ours dataset are ContactPose~\cite{contactpose} and HOD~\cite{huang2022hhor} which also consider a static \ho configuration during the manipulation.  While HOD provides unregistered 3D scans for a subset of the manipulated objects, ContactPose provides groundtruth 3D shapes and poses for both the hand and the object.  This dataset is however limited to objects artifically made textureless, that are equipped with intrusive fiducial markers for motion capture purposes. The hand shape is also obtained after fitting the MANO model.
Besides, we found that some frames are missing in some sequences  leading to discontinuities in \ho motion during manipulation and preventing the use of a video-based approach. 
Our \Ours dataset offers more variety in terms of object appearance and grasp types (see Fig.~\ref{fig:mosaic_orbit}) and, importantly, it is the first dataset that provides  real ground-truth 3D shape for both the hand and the object. 
 We provide a comparison of \Ours  to the most relevant and widely used hand-object interaction datasets in Table \ref{tab:compare_datasets}. 
 
\noindent \textbf{Hand-Object Reconstruction} 
from a single RGB image or from a monocular video is an extremely difficult task due to hand-object mutual occlusions, complex hand-object motion and variability in object shapes. That is why earlier work~\cite{tzionas_capturing_2014,tzionas_capturing_2016,zhang21,ballan2012motion,oikonomidis2011full} considered RGB-D  or multi-view inputs.
Recent works on joint  \ho reconstruction from monocular RGB images have achieved impressive results. These works can be generally categorized into parametric hand model-based methods~\cite{hasson_learning_2019, romero_embodied_2017, cao_reconstructing_2021, liu_semi-supervised_2021, kokic_learning_2019, vedaldi_html_2020} that assume a known object template (or category~\cite{kokic_learning_2019, oberweger2019generalized,hampali2021handsformer}) and implicit representation-based methods~\cite{karunratanakul_grasping_2020}, or a combination of both~\cite{ye_whats_2022,chen_alignsdf_2022}. While~\cite{ye_whats_2022} assumes known 3D templates and obtain both hand and object poses from parametric models - using Signed Distance Functions (SDFs) to help reconstruct shape details for both hand and object,  \cite{chen_alignsdf_2022} only uses a parametric model for the hand prior and reconstruct generic hand-held object without knowing their 3D templates. However, the object reconstruction performance is rather poor as it remains unclear how to learn the implicit representations to reconstruct a large variety of object shapes with a single model as observed in~\cite{karunratanakul_grasping_2020}. To achieve reasonable \ho  results in a fully object-agnostic manner, \cite{huang2022hhor}  leverages multiple observations of a HO rigid configuration along a video sequence. The camera motion is recovered using a hand tracker and an implicit neural representation-based method is then employed to reconstruct the SDF and color fields of the hand and object. Similarly to this method, we consider a 2-stage pipeline consisting of a rigid registration followed by MVR and benchmark several baselines for each of these 2 stages. 

Other methods have considered hand-object monocular \rgb  video as input. 
\cite{hasson_leveraging_2020} performs joint \ho reconstruction by leveraging photometric consistency over time while in \cite{hasson_towards_2022}, an optimization approach is used. \cite{liu_semi-supervised_2021} leverages spatial-temporal consistency to select pseudo-labels for self-training. These methods have the biggest caveat of requiring the object template mesh at inference time, which makes the hand-object reconstruction problem a \ho 6DOF pose estimation task. 
We focus on bench-marking object-agnostic methods that can reconstruct any \ho shapes.

\begin{figure*}[t]
\centering
\begin{minipage}{.69\textwidth}
  \centering
  \includegraphics[width=\linewidth]{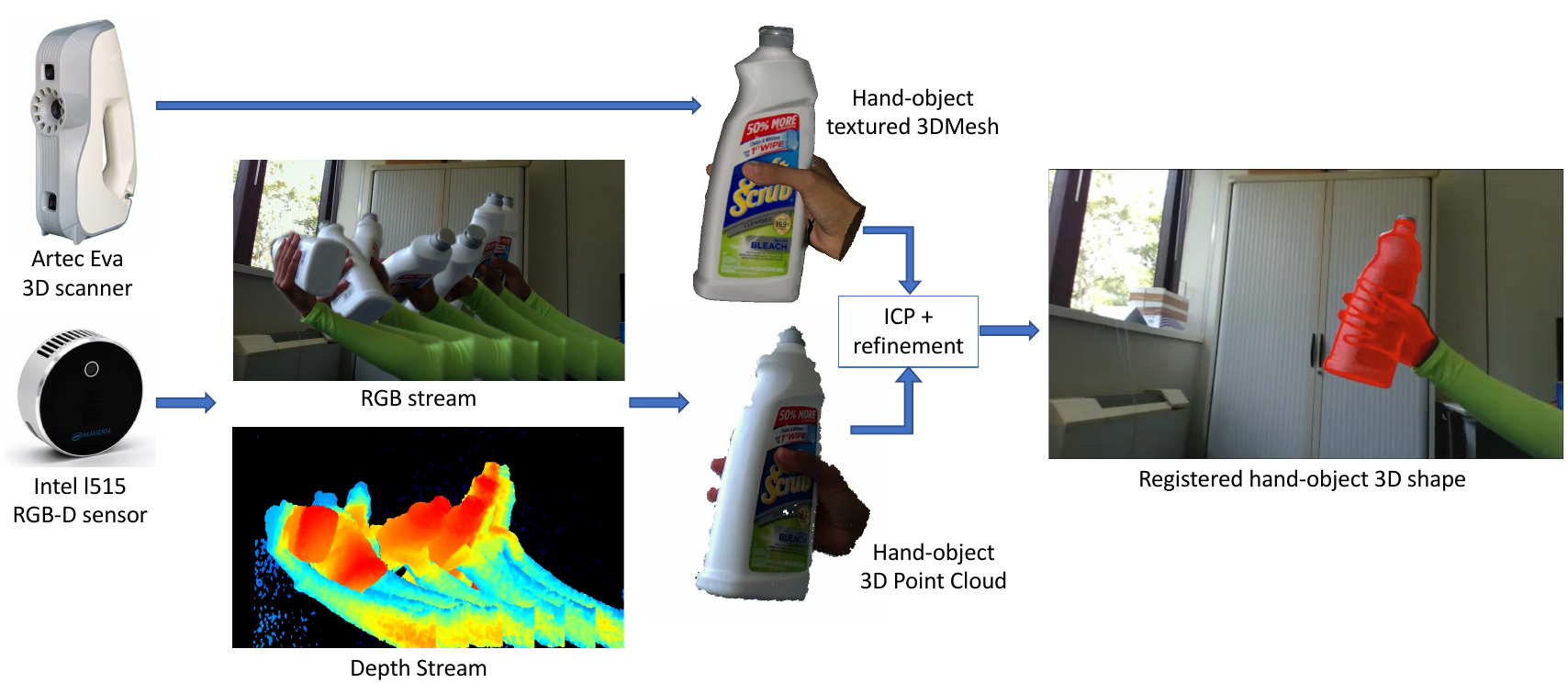} \\[-0.4cm]
  \caption{\textbf{Hand-Object capture and registration pipeline.} 
We capture a video sequence of a hand holding rigidly an object and moving in front of an RGB-D camera, and we automatically segment the hand-object system in the images. We reconstruct a precise textured mesh of the hand-object in the exact same pose using an off-the-shelf 3D scanner and register this mesh to each frame to provide ground-truth annotations.}
  \label{fig:overview_orbit}
\end{minipage}%
\hfill
\begin{minipage}{.29\textwidth}
  \centering
  \includegraphics[width=\linewidth]{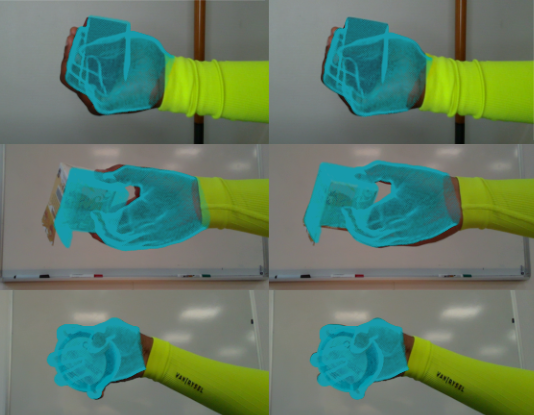}
  \\[-0.35cm]
  \caption{\textbf{Mesh registration procedure for data annotation.} Left: the pose of the ground truth HO mesh (light blue) is initialized through ICP registration with the segmented depth map. Right: it is then refined with differentiable rendering and temporal smoothness priors.  
} 
\label{fig:camrefine} 
\end{minipage}

\vspace{-0.3cm}

\end{figure*}

\section{The \Ours dataset}
\label{sec:dconstruct}

In this section, we detail the collection procedure in \cref{sub:collection} and the data annotation in \cref{sub:groundtruth} (see Fig.~\ref{fig:overview_orbit} for an overview) while \cref{subsec:mano} details how  \gt scans are further annotated with hand-object information.

\subsection{Dataset collection}
\label{sub:collection}
We instruct the subject to grasp one object according to different use cases: either a \textit{power-grasp}, \ie, holding the object strongly with all fingers, a \textit{use-grasp}, \ie, holding the object as if the object was going to be used or a \textit{handover-grasp}, \ie, holding the object as if the intent was to give it to someone else. We then record a video with an RGB\nobreakdash-D monocular camera of the subject showing every part of the hand-object grasp. In order to ease hand-object segmentation from the arm, which is not the focus of our dataset, the subject is wearing a distinctive sleeve and no other human parts are visible in the video.
Once the video is captured, we ask the subject to maintain the same grasp and  capture the shape of the \ho configuration using a sub-millimeter  precision scanner.~\cref{fig:mosaic_orbit} shows several captured textured meshes, highlighting the diversity of objects and grasps.

\noindent \textbf{Hardware details.}
We acquire the videos using a single Intel RealSense L515 \rgbd  camera~\cite{intel_l515}, and we capture the \gt \ho shapes with a Artec Eva 3D scanner~\cite{artec_eva}.
The camera is calibrated in a pre-processing step and is used to capture both depth and \rgb streams at a rate up to 30fps and 1280 x 720 resolution. We process the \rgb and depth streams to perform pixel alignment and temporal synchronization. 
We use the software provided by the supplier for obtaining an accurate shape from the scans.

\noindent \textbf{Dataset statistics.}
We collect 96 sequences from 15 different subjects holding 42 different objects from everyday life, with various sizes and shapes. The subjects reflect diversity in gender, color, and hand shape.
The different grasp types (power-grasp, use-grasp, handover-grasp) are evenly represented. Each video sequence lasts an average of 48 seconds. This represents a total of 87,540 frames.

\subsection{Ground-truth HO 3D shape annotation}
\label{sub:groundtruth}

We now detail how we obtain \ho segmentation in the RGB images and \gt rigid transformation, \ie the alignment between each frame and the 3D mesh obtained from the scanner, allowing its reprojection onto the image.

\noindent \textbf{Segmentation.}
We first segment the foreground, \eg \ho pixels by thresholding the depth values from the input RGB\nobreakdash-D stream. This process segments out the wrist and the object, but also the arm which we want to ignore, since it is out of the scope of this work, and it violates the rigidity assumption. We then segment the arm part by thresholding \rgb pixels values based on the color of the sleeve. Finally, we combine these two masks to obtain the \ho masks 
which can be applied on both the \rgb frames as well as the depth values, that we express as back-projected 3D point clouds.

\noindent \textbf{Rigid transformation from scanned mesh to each frame.} 
For each video, we align all the frames to the scanned \gt mesh. 
The first step of this alignment consists in performing a robust rigid Iterative Closest Point (ICP)~\cite{zhang2021fast} between the \gt mesh and the aforementioned masked depth point clouds. 
We manually 3D align to initialize the first frame of each sequence and then automatically align the remaining frames using the previous result as initialization for the next one,
to obtain initially aligned poses $\{R_i|t_i\}\in SE(3)$, denoting rotations and translations respectively.  We found that such an alignment is already quite satisfactory but some outliers remain, due to sensor noise or invalid local minima of the ICP. We thus  refine these aligned poses via a differentiable rendering pipeline that we detail in the following.

For each sequence, let $I_i, i \in \{1..N\}$ denote the $N$ input frames
of resolution $H \times W$, $S_i$ be the ground-truth segmentations at the same resolution and $\mathcal{M}$ the \gt mesh. 
This mesh is  associated with appearance information acquired from the sensor such that we can render it onto the image planes in a differentiable manner.
Our objective is to refine the camera poses $\{R'_i|t'_i\} = \{ R_i \text{orth}(R_i^{corr}) | t_i+t_i^{corr} \}$ such that the projection of the colored mesh $\mathcal{P}(\mathcal{M},\{R'_i|t'_i\}) $ matches the \rgb observations for each frame. 
We express the pose corrections as offsets over the ICP results. And we parametrize the rotation corrections $R_i^{corr}$ as $2 \times 3 $ matrices, that we orthonormalize with the Gram-Schmidt process $\text{orth}()$ to be rotation matrices, following~\cite{roma}.

More formally,  we minimize a masked Mean Square Error (MSE) between rendered image $\hat I_i$ and  observations:
\begin{equation}
    \mathcal{L}_{RGB} = \sum_i^N \sum_p^{H \times W} S_i(p).|| \hat I_i(p) - I_i(p) ||^2. 
\end{equation}

This loss alone does not converge for sequences where the RGB information is ambiguous. Thus, we add two regularization terms following two assumptions. We assume the consecutive rotations and translations to be smooth, thus we add a smoothing term $\mathcal{L}_{Smooth} = \mathcal{L}_{t} + \mathcal{L}_{R}$ as a combination of two functions that minimize the discrete Laplace operator of transformations, one for rotations $\mathcal{L}_{R}$ in degrees and one for translations $\mathcal{L}_{t}$ in centimetres:
\begin{equation}
    \mathcal{L}_{t} = \sum_{i=1}^{N-1} \frac{|| 2{t'}_i - \sg({t'}_{i-1}+{t'}_{i+1}) ||}{2N},
\end{equation}
\begin{equation}
\mathcal{L}_{R} = \sum_{i=1}^{N-1} \frac{\angle{ \left( \sg(R'_{i-1}), R'_i \right)} + \angle{\left( R'_i, \sg(R'_{i+1}) \right)} }{2N},
\end{equation} 
where $\sg$ is the stop-gradient operator  and $\angle$ returns the angle between two rotations.
$\sg$ is needed to prevent collapsing to unique R and T values in our auto-differentiating framework. These smoothing terms forbid camera transformations that violate the motion smoothness assumption. To incentivize the pose corrections to be small, we add a weight decay regularization term formulated as follows:
\begin{equation}
    \mathcal{L}_{wd} = \sum_i \| R_i^{corr} - I \|^2 + \|T_i^{corr}\|^2
\end{equation} where $I$ denotes the identity rotation. Finally, the final loss we optimize is expressed as:
\begin{equation}
 \mathcal{L} = \mathcal{L}_{RGB} + \lambda_{Smooth} \mathcal{L}_{Smooth} + \lambda_{wd} \mathcal{L}_{wd}
\end{equation}

We did not include a loss for the depth information as it would have been computationally demanding. We considered that the ICP-alignment already provided a signal from the depth, that is included in the current formulation in  $\mathcal{L}_{wd}$.
We model the \gt geometry in the form of a sparse voxel grid structure in the differentiable rendering framework of~\cite{plenoxels}, each voxel close to the \gt mesh having a high opacity. Each non-zero voxel is equipped with appearance information initialized from $\mathcal{M}$. 
As the appearance of $\mathcal{M}$ was obtained using a scanner, it does not correspond exactly to the \rgb observations, so we need to compensate for the appearance to account for sensor-dependent information. We thus optimize for both the camera poses offsets and the appearance of the \gt mesh. Please refer to Appendix A.3 for optimization details.

The effects of this camera refinement procedure are shown in~\cref{fig:camrefine}. Thin structures can hardly be correctly aligned via ICP as only very few pixels provide depth information on those regions. In contrast, the \rgb based refinement along with the smoothing components help annotate more accurate poses.  After manual verification, we managed to improve the annotated poses for $47$ out of the $96$ sequences both quantitatively in terms of $\mathcal{L}_{RGB}$ and qualitatively. The remaining $49$ sequences were already very accurate and the optimization did not help in this case. 

\subsection{Parametric Model Annotations}
\label{subsec:mano}
\begin{figure}
    \centering
    \includegraphics[width=\linewidth]{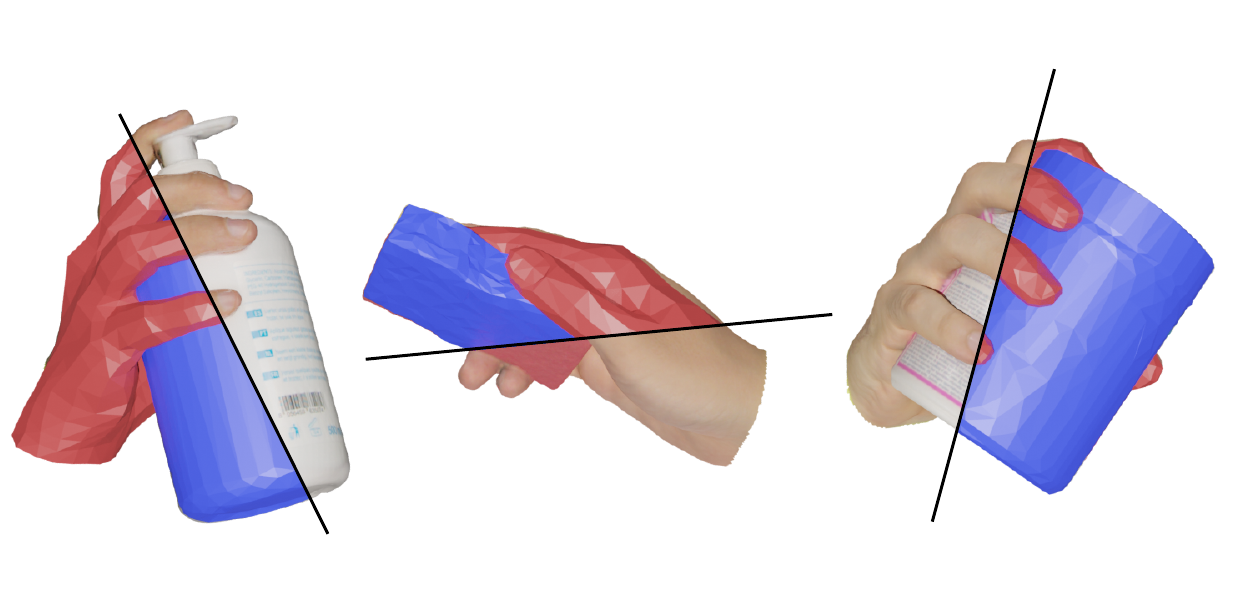}
    \vspace{-0.6cm}
    \caption{\textbf{Hand and Object 3D model annotation.} Partial overlay of the MANO hand model (in red) and a decimated object mesh (in blue) registered to the textured hand-object 3D scan for different sequences of \Ours.    
    }
    \label{fig:mano_annotations}

\vspace{-0.3cm}
    
\end{figure}

For each sequence, we also provide semantic information regarding the depicted grasp.
For that purpose, we captured textured 3D scans of the objects alone that we register together with the MANO hand model~\cite{romero_embodied_2017} to the \ho meshes. This provides pose and shape annotations for both the hand and the object independently, as shown in~\cref{fig:mano_annotations}. 
This additional information could prove useful for other tasks such as detailed grasp analysis, HRI-related tasks or even hand-object pose estimation although out of the scope of this paper. Importantly, the \gt MANO kinematic poses will allow us to benchmark hand pose estimation methods employed to estimate the rigid transformation.

Our three steps registration process is semi-automatic. First, we manually estimate the pose of the object by roughly aligning its mesh to the \ho mesh. Second, we estimate MANO hand pose and shape parameters that minimize the squared distance error between manually annotated 3D keypoints on the \ho mesh and corresponding MANO vertices. We use L-BFGS optimizer and the differentiable MANO layer~\cite{hasson_learning_2019}. Third, we refine MANO parameters and object pose to obtain a precise registration, by minimizing:
\begin{equation}
    \label{eq:registration_objective}
    \cfrac{1}{\vert \mathcal{HO} \vert} \sum_{x \in \mathcal{HO}} \min \left( d(x, \mathcal{O}), d(x, \mathcal{H}) \right),
\end{equation}
consisting in the mean \emph{distance} of each point $x$ on the mesh $\mathcal{HO}$ to the closest point on the hand mesh or the object mesh (denoted respectively $\mathcal{H}$ and $\mathcal{O}$). We define the \emph{distance} between a point $x$ of 3D normal $n_x$ and a mesh $\mathcal{M}$ as:
\begin{equation}
    d(x, \mathcal{M}) \triangleq \|x - p\|^2 + \lambda \|n_x - n_{p}\|^2,
\end{equation}
where $p$ is the point on $\mathcal{M}$ closest to $x$, and where $n_p$ denotes its 3D normal.
We choose $\lambda=1mm^2$ in practice, and sample uniformly $|\mathcal{HO}| = 30k$ points on the \ho mesh to evaluate \cref{eq:registration_objective}. We obtain a sub-millimetre residual error after optimization. 
We provide qualitative visualizations of this registration in~\cref{fig:mano_annotations} and in the project \href{https://europe.naverlabs.com/research/showme}{website} video.

\begin{table*}[!t]
	\centering
	\resizebox{0.85\linewidth}{!}{%
		\begin{tabular}{ll|ccc|ccc}
			\specialrule{.1em}{.05em}{.05em}
			& \multirow{2}{*}{Method} & \multicolumn{3}{c|}{Hand pose} & \multicolumn{3}{c}{Rigid transformation} \\
			 &  & MPJPE $\downarrow$ & PA-MPJPE $\downarrow$ & PCK $\uparrow$ & Rot error $\downarrow$  & Trans error $\downarrow$ & Det. rate (\%) $\uparrow$ \\
			\midrule
			\multirow{2}{*}{image-based}
			& Minimal Hand~\cite{zhou2020monocular} &  85.4	& 38.1 & 10.9  & - & - & -\\ 
                & Frankmocap~\cite{frankmocap} &  39.3	& 14.9	& 38.3 & - & - & -\\ 
			& HandOccNet~\cite{park2022handoccnet} &  37.4	& 14.7	& 45.7 & - & - & -\\
                & DOPE ~\cite{weinzaepfel2020dope} &  \bf{26.9} &	\bf{12.4} & \bf{64.6} &	21.0 & 0.17 & 99.0\\
			\midrule
		     
		   \multirow{4}{*}{video-based} & DOPE ~\cite{weinzaepfel2020dope} + fixed hand pose &  \textbf{26.2} &	12.4 & \textbf{69.4} &	21.5 &	0.16 & 99.0\\
		    & DOPE ~\cite{weinzaepfel2020dope} + median filtering & \bf{26.2} & 12.4 &	\textbf{69.4} & 21.3 & 0.15 & \bf{100} \\
		    & DOPE ~\cite{weinzaepfel2020dope} + PoseBERT ~\cite{baradel2022posebert} &  27.3 & \textbf{12.3} &	58.4 & 20.6 & 0.15 & \bf{100}\\
			
			& COLMAP ~\cite{schoenberger2016sfm} &  - &	- &	- &	\textbf{14.6} &	\textbf{0.06} & 78.2\\
            \specialrule{.1em}{.05em}{.05em} \\
		\end{tabular}%
	}
\vspace{-0.65cm}
\caption{
\textbf{MANO Evaluations: Hand pose estimation and associated rigid transformation estimation.}
The MPJPE and PA-MPJPE are reported in mm.
We use a threshold of 30mm for the PCK.
The `Rot. error' is the geodesic distance expressed in degree with the ground-truth rigid transformation.
The `Trans error' is the MSE.
}
\label{tab:hand_pose}

\vspace{-0.3cm}

\end{table*}

\section{Two-stage reconstruction pipeline}
\label{sec:pipeline}
To reconstruct the \ho from an RGB video, We use a 2-stage pipeline in \cref{fig:overviewTask}: estimating the rigid transformations of the \ho in the sequence (\cref{sub:rigid}) and MVR (\cref{sub:mvs}).
\subsection{Rigid transformation estimation}
\label{sub:rigid}
We evaluate two methods for estimating the rigid transformation of the \ho between frames, either using standard generic SfM toolbox, or using the hand pose as a proxy. 

\noindent \textbf{Rigid transformation from a SfM toolbox.}
We run COLMAP~\cite{schoenberger2016sfm} -- SfM software recognized for its robustness and efficiency -- to estimate the pose of the camera with respect to the \ho system across video frames. We ignore background keypoints using the silhouettes information.
\noindent \textbf{Rigid transformation from hand pose estimation.}
In our particular setup, we can also measure the rigid transformation 
by estimating the \ho pose. As in~\cite{huang2022hhor}, we assume the object to be unknown and we focus on the hand keypoints.
We first run an off-the-shelf 2D-3D hand pose estimator, and estimate the rigid transformation  between frames by computing the relative transformation of the hand 3D keypoints. As these are centered around the wrist, while 2D keypoints are estimated in the pixel space, we first run a PnP algorithm to obtain 3D keypoints in the scene.
Then, we estimate the rigid transformation, \ie camera poses, between frames via Procrustes alignment. 

\subsection{Reconstruction from multiple observations}
\label{sub:mvs}

\noindent \textbf{Reconstruction from robust visual hulls (VH).}
First, we consider the silhouette-based  formulation from~\cite{leroyvh} as a baseline for reconstruction, using \gt silhouettes. Following their notation, we set $\alpha=N/8$ and $\beta=N/4$. 

\noindent \textbf{Reconstruction with fast differentiable rendering (FDR).}
We also benchmark the recent 
method from~\cite{toussaint2022fast}.
They propose a coarse-to-fine differentiable rendering method,  targeted at multiview surface capture problems.

\noindent \textbf{Reconstruction with neural implicit surfaces}.  We finally consider the more advanced method proposed in HHOR~\cite{huang2022hhor} that combines NeuS~\cite{wang2021neus}, a NeRF representation where the density radiance field is replaced with a Signed Distance Field (SDF),  with semantic-guided ray sampling (to focus more on the object) and a camera refinement stage. This step simultaneously optimizes SDF and camera poses to compensate for imprecise estimations. 

\section{Experimental results}
\label{sec:expe}
We now evaluate the 2 stages of the pipeline, namely rigid registration (Section~\ref{sub:evalrigid}) and MVR (Section~\ref{sub:evalrecons}).

\subsection{Rigid transformation estimation evaluation}
\label{sub:evalrigid}

We report results for estimating the rigid transformations either from hand poses or from COLMAP in Tab~\ref{tab:hand_pose}. 
As the performance for the hand-based method is likely correlated with hand pose accuracy, we also evaluate hand 3D pose estimation for 4 different image-based methods: (i) Minimal Hand~\cite{zhou2020monocular} an easy to use real-time system, 
(ii) FrankMocap~\cite{frankmocap}, used in IHOI~\cite{ye_whats_2022} and HHOR~\cite{huang2022hhor},   
(iii) the recent HandOccNet~\cite{park2022handoccnet} and (iv) the hand module of DOPE~\cite{weinzaepfel2020dope} which proved to perform well under hand-object interactions~\cite{ArmaganGBHRZXCZ20}. We found that DOPE outperforms the other methods by a large margin and selected it as hand pose estimator.

We also investigate three methods to further smooth the per-frame DOPE predictions: (i) Exploiting the rigid motion assumption, by computing a median pose resulting from an aggregation of all hand poses across the sequence. (ii) By applying a median filter on pose sequences, with a sliding window of 5 frames. (iii) Using PoseBERT \cite{baradel2022posebert} a transformer module for smoothing 3D pose sequences. We found simple baselines (i) and (ii) to perform better.

\begin{table*}[!t]
	\centering
	\resizebox{0.85\textwidth}{!}{%
		\begin{tabular}{ll|cccccc}
			\specialrule{.1em}{.05em}{.05em}
			
			Rigid & Recon. & Rec. rate &  Acc.$^\dagger$ & Comp.$^\dagger$ & Acc. ratio  & Comp. ratio  & Fscore  \\ 
			Transform & Method &  (\%) $\uparrow$ & (cm) $\downarrow$ &  (cm) $\downarrow$ &  @5mm (\%) $\uparrow$ & @5mm (\%) $\uparrow$ & @5mm (\%) $\uparrow$ \\ 
			\midrule
            \color{red}GT  &  IHOI ~\cite{ye_whats_2022} & 87.3 &  0.79  & 1.34 & 41.7 & 37.8 & 39.3 \\ 
                \midrule
			{\color{red} GT} & \vh~\cite{leroyvh} & 93.7 & 0.42 &	0.65 &	67.3 & 	61.6 & 	63.6 \\ 
			{\color{red} GT} & \briac~\cite{toussaint2022fast} &  95.8 & 0.35	& 0.49		&  75.8 	& 	72.0 	& 	73.5	\\ 
                {\color{red} GT} & \sigg~\cite{huang2022hhor} &  \textbf{98.9} & \bf{0.34}	& \bf{0.31}		&  \bf{81.0} 	& 	\bf{83.7} 	& 	\bf{82.2}	\\ 
		
			\midrule  
                DOPE~\cite{weinzaepfel2020dope} &  \briac~\cite{toussaint2022fast}  & \textbf{92.7} & 1.02 &	3.18	&31.7 &	15.7	&20.0 \\ 
		    COLMAP~\cite{schoenberger2016sfm} &  \briac~\cite{toussaint2022fast} & 76.0 & \textbf{0.64} & 0.79 & 39.3 & 36.2  & 37.6 \\ 
                COLMAP~\cite{schoenberger2016sfm} &  \sigg~\cite{huang2022hhor} & 72.9 & 0.65 & \textbf{0.73} & \textbf{53.7} & \textbf{55.2}  & \textbf{54.2} \\ 
            \specialrule{.1em}{.05em}{.05em} \\
		\end{tabular}%
	}
\vspace{-0.75cm}
\caption{
\textbf{Hand-object reconstruction evaluation} using either ground-truth rigid transformations or estimated ones.
$^\dagger$ means that the metrics are obtained by computing on the reconstructed mesh only, the failing ones are not taken into account, making direct comparison between different methods unfair. 
DOPE refers to the variant `DOPE + fixed hand pose' from Tab \ref{tab:hand_pose}.
}
\label{tab:recon}
\vspace{-0.28cm}
\end{table*}
We found that better hand pose estimations tend to lead to better rigid transformations but COLMAP performs the best. 
 However, it yields a lower detection rate compared to its hand pose counterpart (which always provides an estimation), requires accurate segmentation and recovers the camera poses up to an unknown scale factor. Hand-based poses naturally embed a rough scale information and the resulting reconstructions have a similar scale to that of \gt meshes, which is an interesting property.
 
\subsection{Hand-object 3D Reconstruction evaluation}
\label{sub:evalrecons}

In~\cref{tab:recon}, we report accuracy (acc), completeness (comp), and Fscore for different reconstruction methods after Procrustes in rotation, translation and scale to the \gt scans. First, we evaluate the performance of IHOI~\cite{ye_whats_2022}, a recent single-image \ho reconstruction method.  We use the annotated MANO joints for alignment, which is thus near-perfect. This explains the overall good results despite severe artefacts in the reconstructions (see Appendix B.3). On the other hand, these results show that a strong hand prior helps for this challenging task. The reconstruction rate reported in the table is expressed frame-wise for this method.

\begin{figure}
 \centering 

\vspace{-1mm}

 \includegraphics[width=1.0\linewidth]{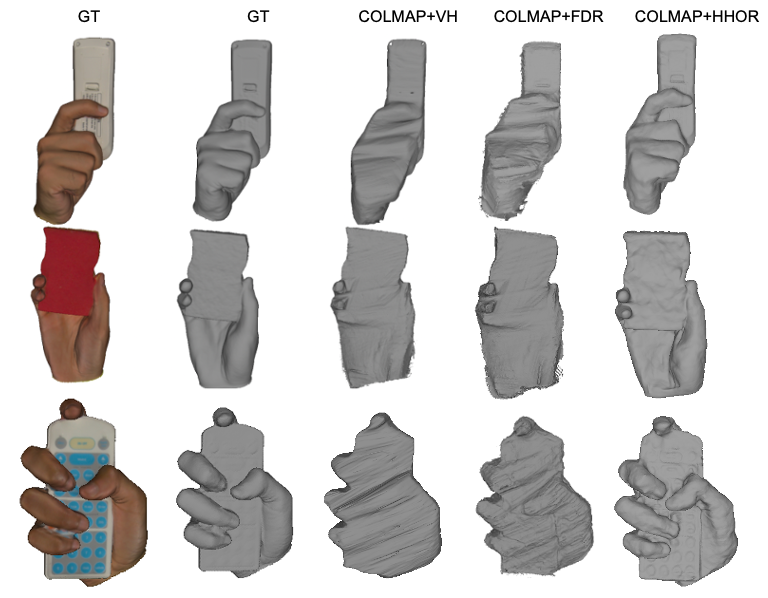} \\[-0.4cm]
 \caption{
 \textbf{Qualitative reconstruction results.} 
 } 
 \label{fig:quali_final}

 \vspace{-0.43cm}
 
 \end{figure}

Using \gt rigid transforms, all 3 reconstruction methods lead to an excellent result (Fscore above 60\% at $5mm$). The recent HHOR method performs better for all metrics. We then evaluate the FDR reconstruction using estimated rigid transforms from either hand keypoints or SfM. The performance drops, \eg from a Fscore @5mm from 73.5\% to 37.6\% using COLMAP, and to 20\% using DOPE. Next, we evaluate HHOR and observed a 16.6\% boost compared to FDR (vs a 9\% boost only when using \gt rigid transforms). The camera pose refinement corrects noisy camera poses from COLMAP at the expense of a much heavier computational cost (1 GPU.day per sequence for HHOR \vs less than a minute for FDR). We show qualitative results in Fig.~\ref{fig:quali_final} and in Appendix B.1. VH cannot reconstruct concavities, \eg, between fingers, while FDR is slightly better. We can appreciate that the shapes reconstructed by HHOR are highly-detailed. Note that HHOR~\cite{huang2022hhor} reported very poor results with COLMAP, justifying the use of FrankMocap to estimate the rigid transforms. However, FrankMocap performs very poorly on our dataset of varied \ho interactions. We posit that the unrealistic close-up fingertip grasps in their HOD dataset allowed accurate hand pose estimates, and it is not the case at all in our setup.

  \noindent \textbf{Detailed analysis.}
Upon careful analysis, we found that COLMAP failed or performed poorly on objects of small size compared to larger-size objects. To corroborate this, we categorize the objects in our dataset to small and larger (\ie,  large and medium) objects and compute reconstruction errors on these two sets of objects. Table   \ref{tab:fdrcolmaprecon} shows the  reconstruction metrics. We observe that COLMAP leads to better results on larger objects while DOPE is better for small objects. For small objects, there may not be sufficient features detected for the matching step which is critical for camera pose estimation by COLMAP.  On the other hand, small objects lead to less hand occlusions and better hand joint estimates, which in turn results in robust rigid-transformation estimation. This strongly emphasizes that a robust hand key points estimator is key for accurate rigid-transformation estimation in the case of small objects with little visual support to perform a standard pose estimation. 

\begin{table}[]
\resizebox{\linewidth}{!}{%
\begin{tabular}{ccccc}
\hline 
  \multirow{2}{*}{method}&object  & Acc. ratio & Comp. ratio & Fscore \\ 
&size & @5mm (\%) $\uparrow$ & @5mm (\%) $\uparrow$ & @5mm (\%) $\uparrow$ \\
\midrule
\multirow{2}{*}{COLMAP+FDR}&small              & 31.78                         & 28.95                          & 30.23                     \\ 
&larger          & \textbf{50.05}                         & \textbf{46.23}                          & \textbf{47.93}                     \\ 
\midrule
\multirow{2}{*}{DOPE+FDR}&small              & \textbf{35.38}                          & \textbf{18.58}                          & \textbf{23.43}                     \\ 
&larger          & 29.44                          & 13.96                          & 17.85                     \\ \hline
\end{tabular}%
}
\\[-0.27cm]
\caption{ \textbf{\ho reconstruction evaluation \vs object size.}}
\label{tab:fdrcolmaprecon}

\vspace{-0.3cm}

\end{table}

\section{Conclusion}
\label{sec:conclusion}

We introduced the \Ours{} dataset to tackle the problem of detailed 3D reconstruction of a hand holding rigidly an unknown object from a monocular video. We then benchmarked several video-based baselines that follow a two-stage pipeline consisting of a rigid registration step followed by a multi-view reconstruction. Even if high-quality \ho 3D reconstructions are obtained in some cases, their quality highly depends on the initial rigid transformation estimates which are difficult to obtain for texture-less objects or heavy hand occlusions. There is still room for improvement regarding the reconstruction quality too and we hope \Ours{} will help foster further research in this direction.

\noindent \textbf{Acknowledgements. }This work was supported in part by
MIAI@Grenoble Alpes (ANR-19-P3IA-0003)

{\small
\bibliographystyle{ieee_fullname}
\bibliography{biblio}
}

\end{document}


\title{SHOWMe: Benchmarking Object-agnostic Hand-Object 3D Reconstruction \\ \textit{Appendix}}

\author{Anilkumar Swamy$^{1,2}$ \hspace{0.1cm} Vincent Leroy$^{1}$ \hspace{0.1cm} Philippe Weinzaepfel$^{1}$ \hspace{0.1cm}  Fabien Baradel$^{1}$ \hspace{0.1cm} Salma Galaaoui$^{1}$ 
\and
\hspace{0.1cm} Romain Brégier$^{1}$ \hspace{0.1cm} Matthieu Armando$^{1}$ \hspace{0.1cm} Jean-Sebastien Franco$^{2}$   \hspace{0.1cm} Grégory Rogez$^{1}$
\and
\textsuperscript{1}\footnotetext[1]{}{}{NAVER LABS Europe} 
\textsuperscript{2}\footnotetext[1]{}{}{Inria centre at the University Grenoble Alpes}\\
}

\twocolumn[{%
\renewcommand\twocolumn[1][]{#1}%
\maketitle
\begin{center}
    \captionsetup{type=figure}
    \includegraphics[width=\linewidth]{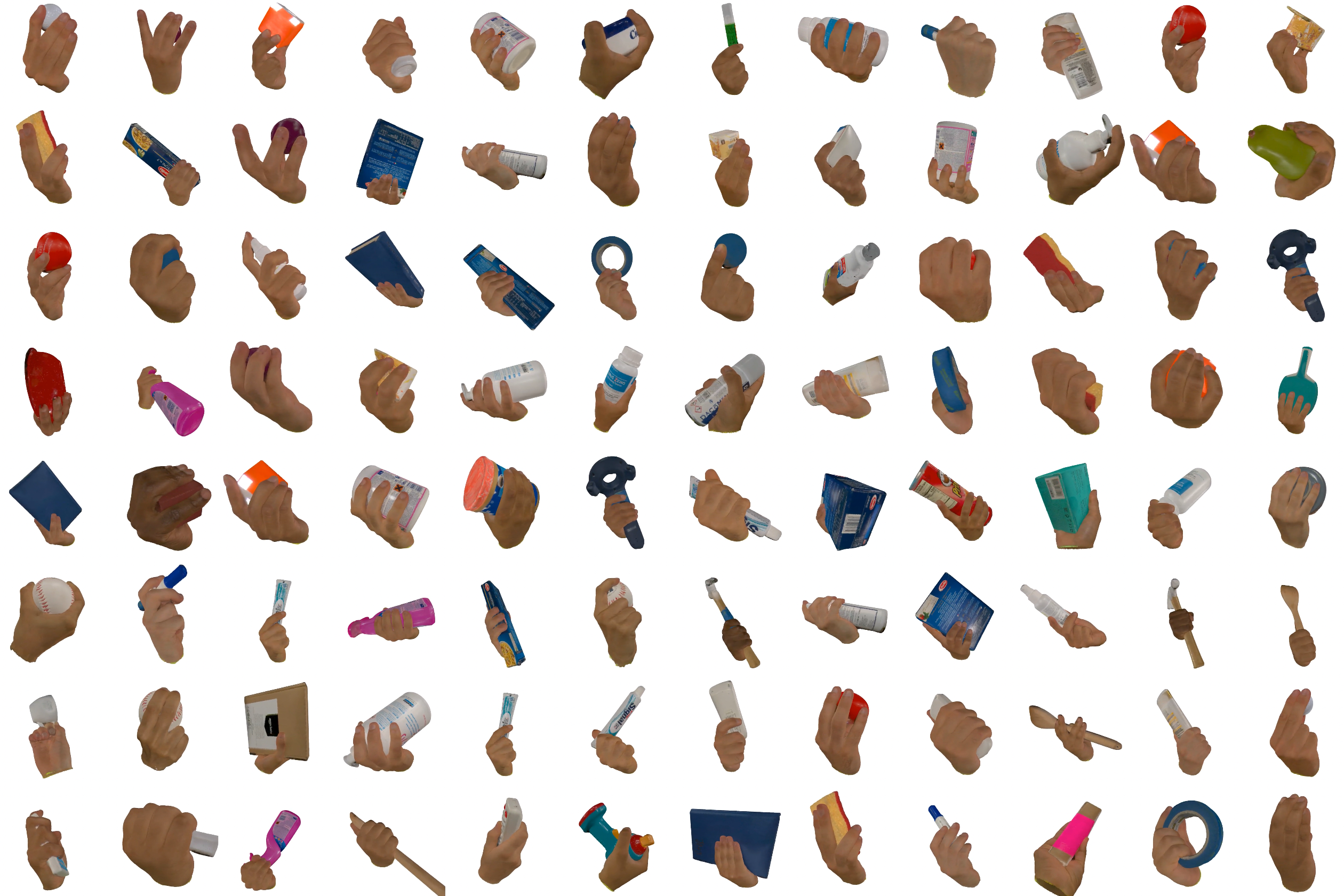}

    
    \captionof{figure}{ \textbf{Snapshot from the video of the SHOWMe dataset} with the 
    ground-truth textured meshes it contains.}
    \label{fig:gt_mesh_mosaic}

    \vspace{0.8cm}
\end{center}
}]

In this Appendix, we provide more details about the introduced SHOWMe dataset in Section~\ref{app:dataset}. 
Section~\ref{app:results} then provides a more detailed evaluation analysis of the results of our two-stage pipeline on the \Ours dataset, as well as the method based on aggregated single-view reconstruction from IHOI~\cite{ye_whats_2022}.

\begin{figure*}
    \centering
    \includegraphics[width=\linewidth]{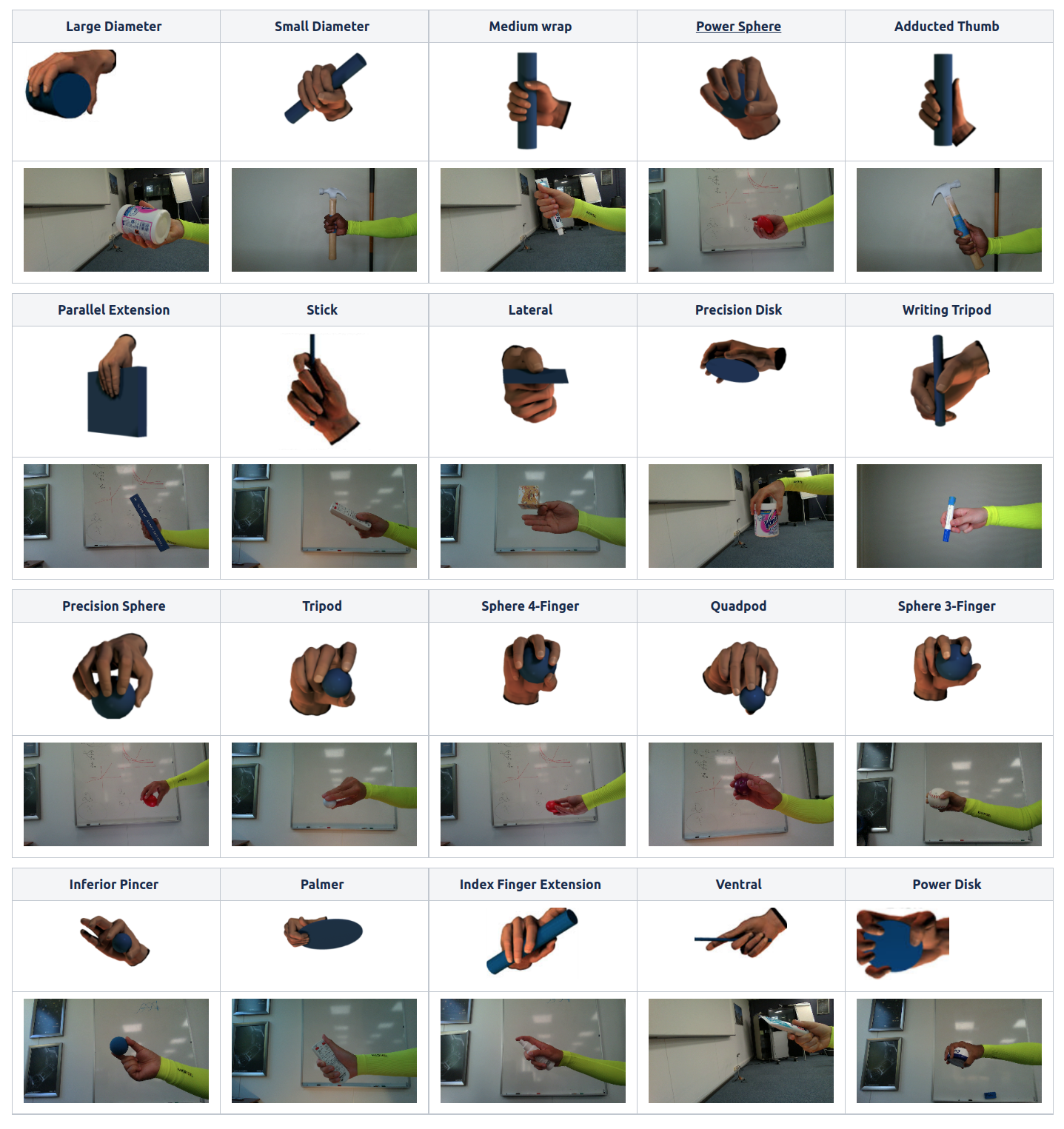}
    \vspace{-0.8cm}
    \caption{\textbf{Grasp categories within \Ours}.}
    \label{fig:grasp_cat}
\end{figure*}

\section{Additional Dataset Information}
\label{app:dataset}

\subsection{Qualitative Visualizations}
\label{subsec:quali}

The attached video, from which we show a snapshot in Figure~\ref{fig:gt_mesh_mosaic}, highlights the ground-truth 3D shapes with textures that our introduced SHOWMe dataset contains. With the scanner accuracy and resolution
up to 0.1mm and 0.2mm respectively, the \#vertices and \#faces depend on the size and shape of the objects. Across
all GT meshes, \#vertices and \#faces are in the range (4K to 263K) and (7.5K to 524K) respectively.

\subsection{Grasp Variability}
\label{subsec:grasp}
\Ours offers a large variability in terms of grasp types as depicted in Figure~\ref{fig:grasp_cat} where 20 of our hand-object interactions can be classified into different classes of the 33-grasp taxonomy introduced in \cite{Feix_33grasps_2009}. 

\subsection{Rigid transformation - Optimisation details}
\label{subsec:optim}
 
As explained in Section 3.2, we refine the camera poses such that the projection of the \gt colored mesh matches the \rgb observations for each frame and minimize a loss expressed in Equation (5).

We implement this optimization in PyTorch~\cite{pytorch}. We represent the scene with a sparse voxel grid of resolution $128^3$, that matches the bounding box of~$\mathcal{M}$. 
We use the Adam optimizer~\cite{adam} for $250$ iterations, with $lr_{rgb}=5.10^{-1}$ and $lr_{cam}=5.10^{-3}$ the learning rates for the appearance and the camera poses respectively. 
Because it is not computationally tractable to fully render thousand of frames at every iteration, we randomly sample $500$ rays from each camera at each iteration. We grid search various $\lambda_{Smooth}$, $\lambda_{wd}$ weighting parameters and keep for each sequence the set of hyperparameters that gave the best (lowest) $\mathcal{L}_{RGB}$ value, considering that this metric is a relevant proxy for the quality of the pose annotations.

\section{Detailed Evaluations of Baselines}
\label{app:results}
In this section, we present in~\cref{subsec:resquali} additional qualitative results of the various baselines.
We then provide in~\cref{subsec:quanti} more detailed and comprehensive experiments on the different steps of these baselines, before discussing single-image reconstruction results in \cref{sub:ihoi}.

\subsection{Qualitative Results}
\label{subsec:resquali}

\begin{figure}
    \centering
    \includegraphics[width=\linewidth]{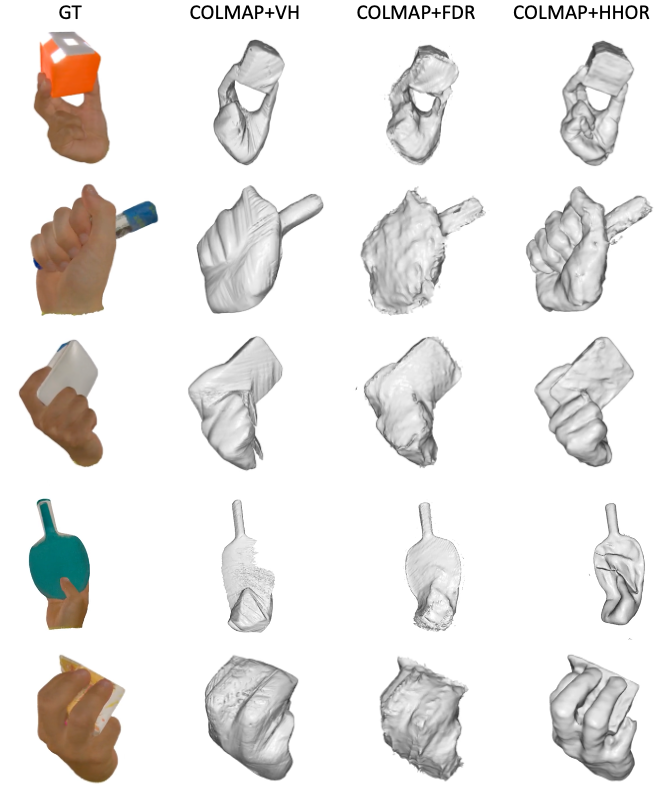}
    
    \vspace{-0.5cm}
    
    \caption{\textbf{Hand-Object Reconstructions} using COLMAP for camera pose estimation followed by VH, FDR and HHOR methods for reconstruction. \textit{From left to right:} \gt, COLMAP+VH, COLMAP+FDR, COLMAP+HHOR. Please see the video for more qualitative results.}
    \label{fig:gt_cVH_cFDR_cHHOR}
\end{figure}

We provide in~\cref{fig:gt_cVH_cFDR_cHHOR} successful reconstructions obtained with VH, FDR and HHOR methods, using camera poses obtained from COLMAP. HHOR recovers much more details for both the hand and the object compared to VH and FDR.

\begin{figure}
    \centering
    \includegraphics[width=\linewidth]{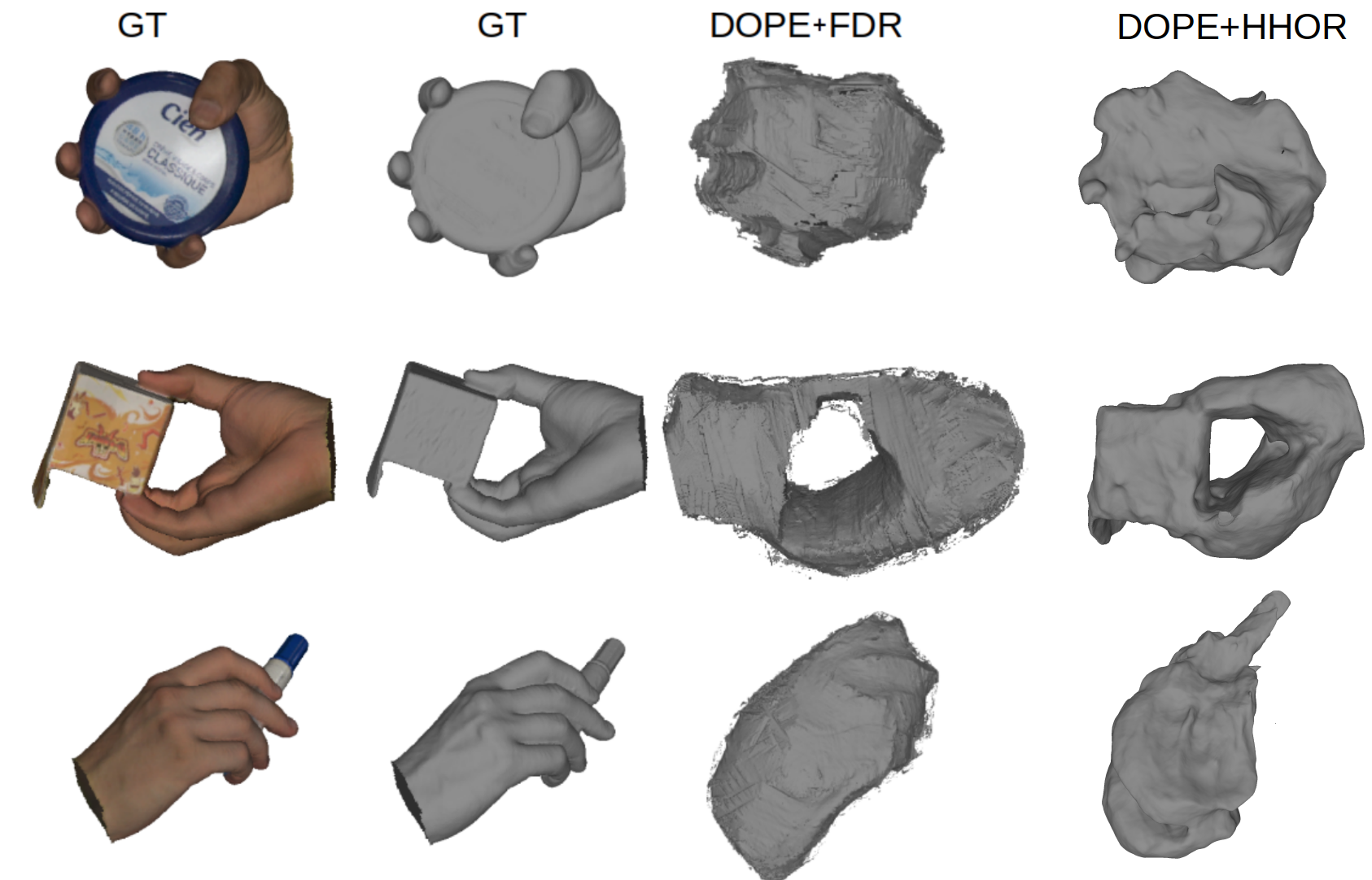}
    
    \vspace{-0.3cm}
    
    \caption{\textbf{Hand-Object Reconstruction using DOPE+FDR and DOPE+HHOR.}For these examples, COLMAP could not estimate any valid camera poses.}
    \label{fig:dope_recon_fdr_hhor}
\end{figure}

\begin{figure}
    \centering
    \includegraphics[width=\linewidth]{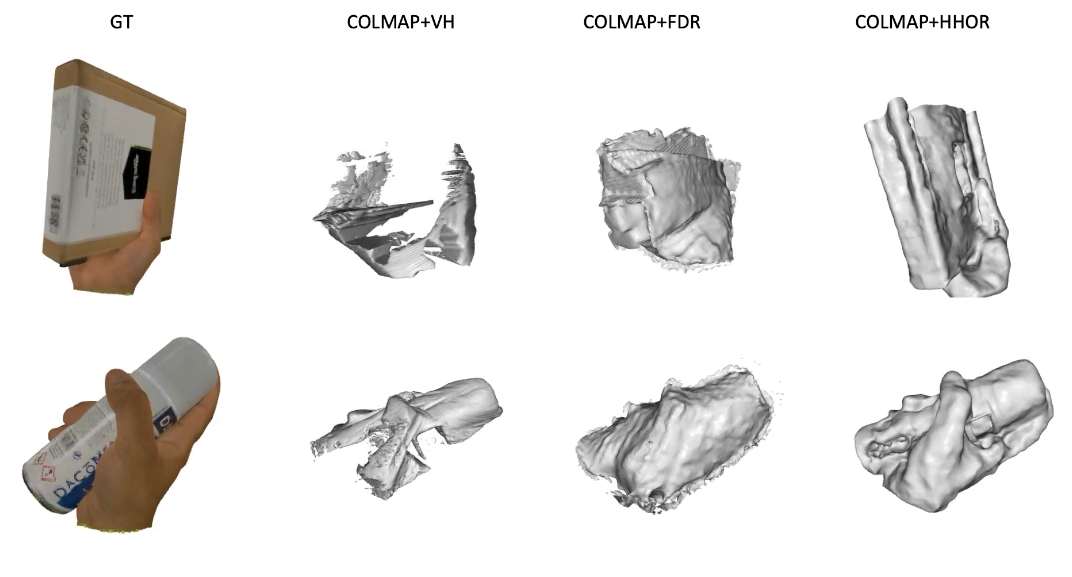}
    
    \vspace{-0.3cm}
    
    \caption{\textbf{Hand-Object Reconstruction failure examples} obtained using COLMAP as pose estimator. The poses estimated are not accurate enough to allow for reasonable reconstruction results.}
    \label{fig:failure_case}
\end{figure}

\subsection{Detailed analysis}
\label{subsec:quanti}

\begin{figure}[ht]
    \centering
    \subfloat[]{\includegraphics[width=0.25\textwidth]{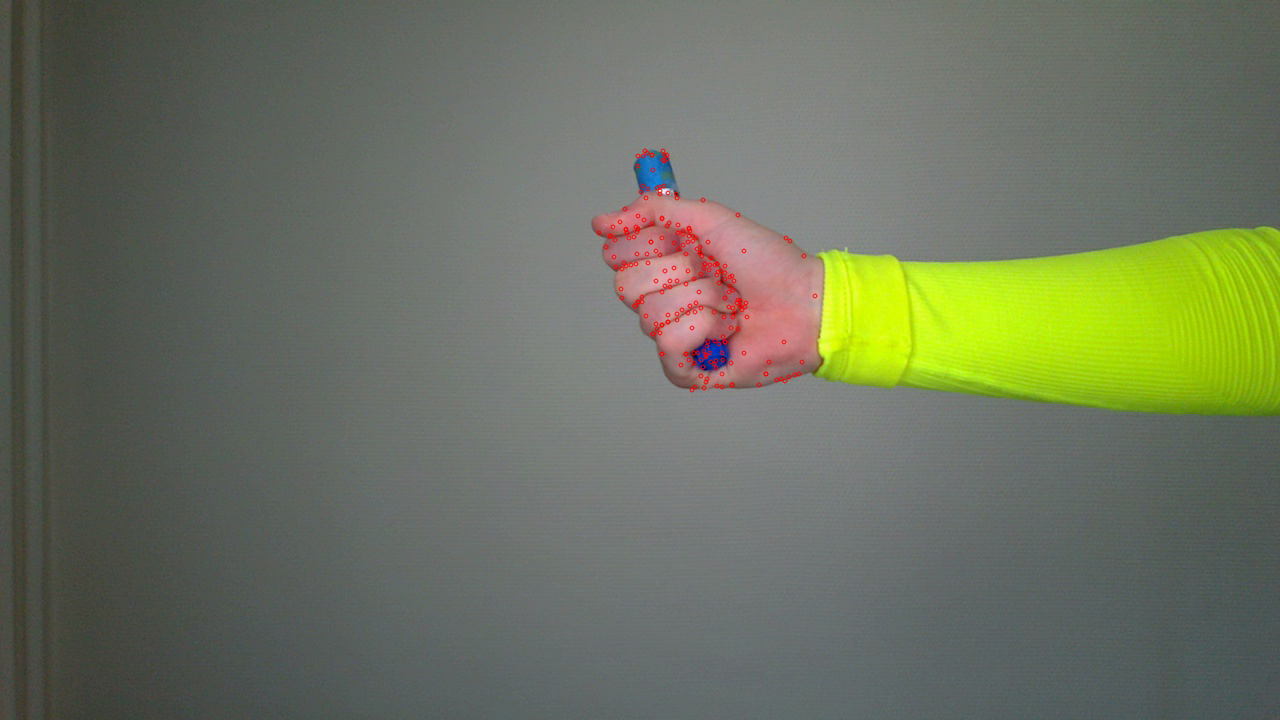}}
    \subfloat[]{\includegraphics[width=0.25\textwidth]{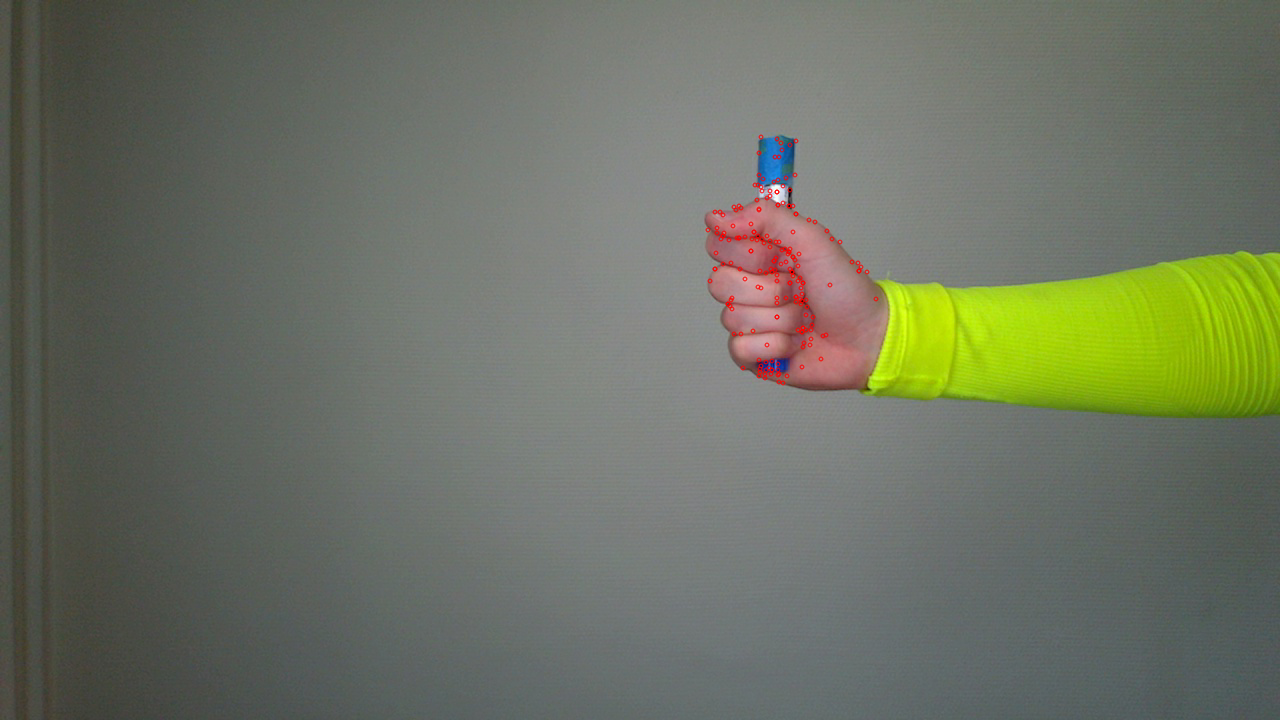}}
    \\
    \subfloat[]{\includegraphics[width=0.5\textwidth]{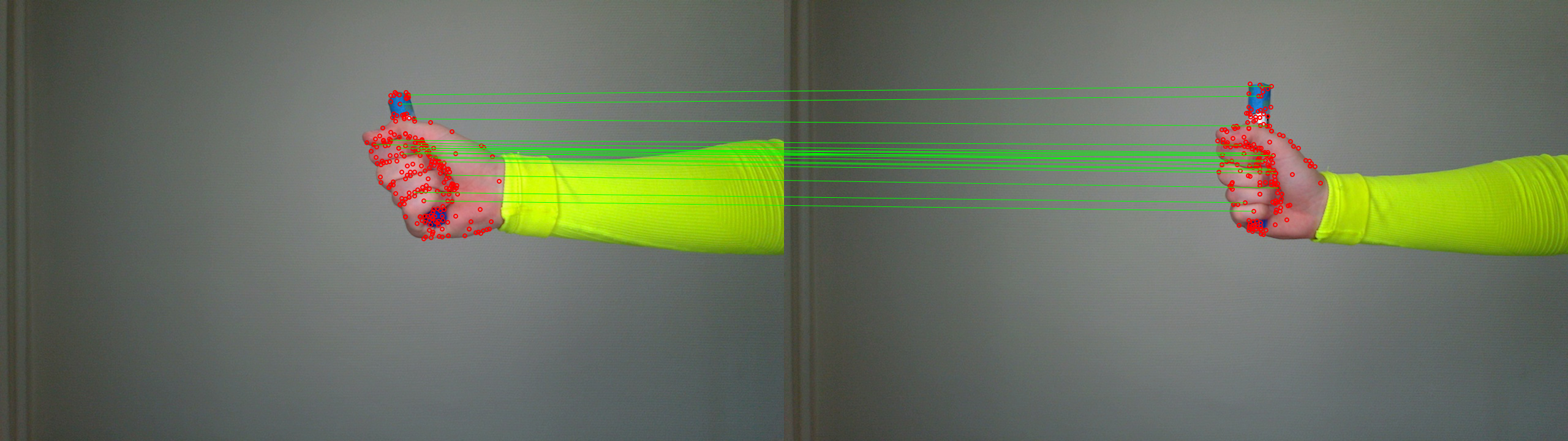}} \\[-0.2cm]
    \caption{\textbf{Visualization of COLMAP detected features} on (a) frame 1 and (b) frame 264. We show the corresponding matched features in (c).}
    \label{fig:colmap_feats}
\end{figure}

In this section, we analyze the success and failure cases of our two baselines used for rigid transformation estimation which plays a crucial role in the final 3D reconstruction quality. COLMAP~\cite{schoenberger2016sfm} camera pose estimation toolbox takes the set of overlapping images of the same object from different viewpoints and estimates the intrinsic and extrinsic camera parameters. This involves 2D images feature detection, extraction, and feature-matching steps. Figure~\ref{fig:colmap_feats} shows the example of detected features in two viewpoints and corresponding matched features. However, we observe that COLMAP fails to estimate the camera poses on roughly 20\% of the sequences. An example of complete failure cases can be seen in Fig.~\ref{fig:dope_recon_fdr_hhor} where COLMAP could not successfully output camera poses. This figure also shows the reconstruction of DOPE+\briac and DOPE +HHOR, and illustrates the robustness of a hand-based camera estimation approach, at the cost of a loss in reconstruction quality. In other cases, camera poses are estimated by COLMAP but are not sufficiently accurate to perform reasonable reconstruction as seen in~\cref{fig:failure_case}.

\noindent \textbf{Object Size Analysis.}
Upon careful analysis in section 5.2, we found that reconstructions based on COLMAP failed or performed poorly on objects of small size compared to larger-size objects while using DOPE led to better performance for
small objects. To corroborate this, we manually label each object as small or large and compute rigid transformation estimation qualities and reconstruction errors on the two sets (small size set and larger size set) of objects.
Table \ref{tab:colmaprelpose_size_newmet_perframe} and \ref{tab:doperelpose_size_newmet_perframe} shows the rigid transformation accuracy averaged across sequences within \Ours for three different level of camera pose precision (\ie, after binning camera estimates with respect to translation and rotation errors). COLMAP shows comparable average performance for both types of objects but a much higher variance in camera pose quality for small objects, meaning that it completely fails for some sequences. On the other hand, DOPE clearly provides more accurate camera estimates for small objects. Using HHOR as a reconstruction method does not compensate for mediocre camera estimates and the same observation can be made (Table~\ref{tab:hhordoperecon_size}).

\noindent \textbf{Object Texture Analysis.} We also analyze the success and failure cases of our baselines used for rigid transformation estimation and 3D reconstruction with respect to the texture quality of the objects in the sequence. 
We have empirically observed that COLMAP camera pose estimation quality reduces drastically for less textured objects, \ie, with a rather uniform texture.
In order to verify this observation, we performed an experiment where we manually categorize all the hand-object sequences into two types "\textit{textured}" and "\textit{less-textured}" objects and evaluate the  rigid transformations estimated with COLMAP in Table~\ref{tab:colmaprelpose_tex_newmet}.
COLMAP shows comparable average performance for both types of objects but a much higher variance in camera pose quality for less-textured objects, meaning that it completely fails for some sequences.
We also evaluate the reconstruction baseline COLMAP+HHOR  for the two sets and found that acc. ratio, comp. ratio and Fscore all degrade by roughly 15\%  as shown in Table \ref{tab:hhorcolmaprecon_tex}.

\begin{figure}
     \centering
    \subfloat{\includegraphics[width=0.235\textwidth]{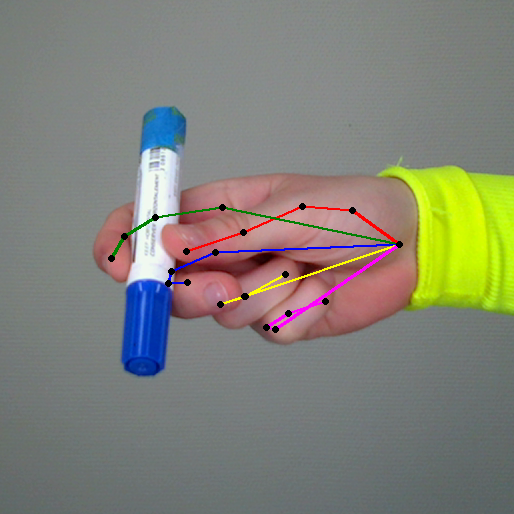}} \hfill
    \subfloat{\includegraphics[width=0.235\textwidth]{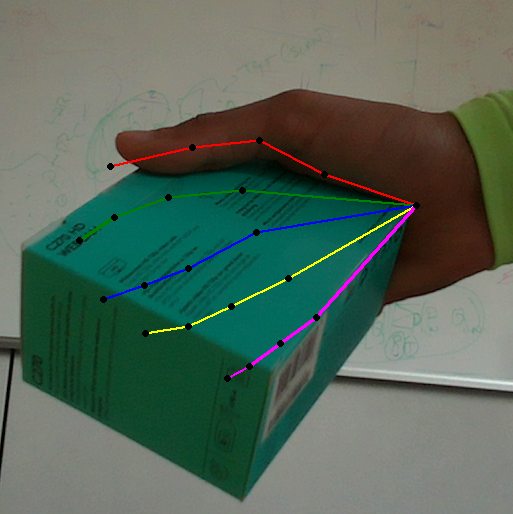}}\\
    
    \subfloat{\includegraphics[width=0.235\textwidth]{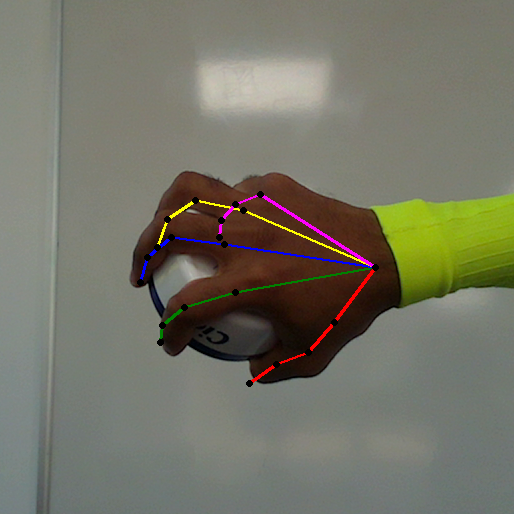}} \hfill
    \subfloat{\includegraphics[width=0.235\textwidth]{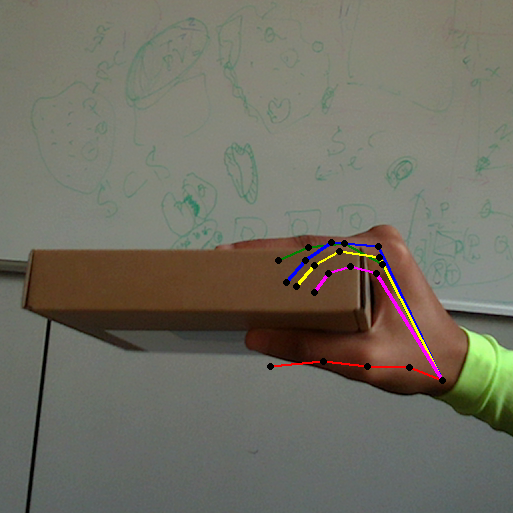}}\\
    
    \subfloat{\includegraphics[width=0.235\textwidth]{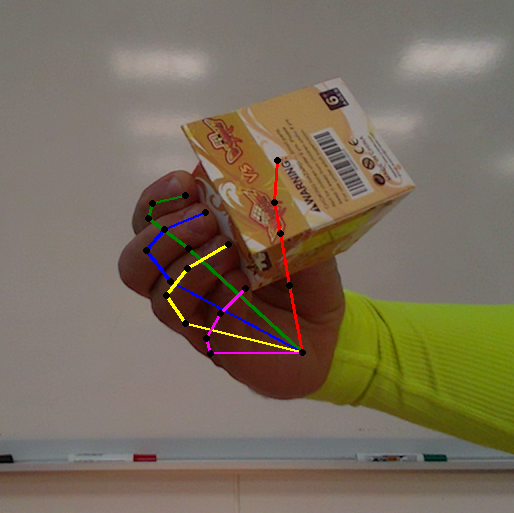}} \hfill
    \subfloat{\includegraphics[width=0.235\textwidth]{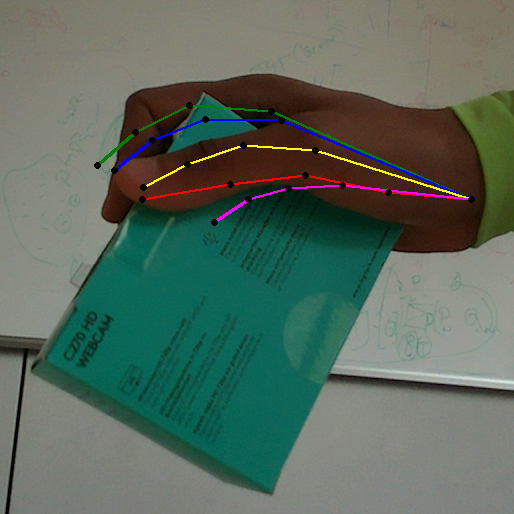}}\\

    \subfloat{\includegraphics[width=0.235\textwidth]{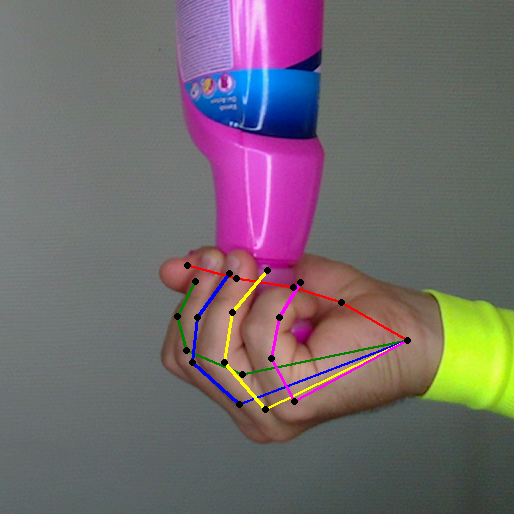}} \hfill
    \subfloat{\includegraphics[width=0.235\textwidth]{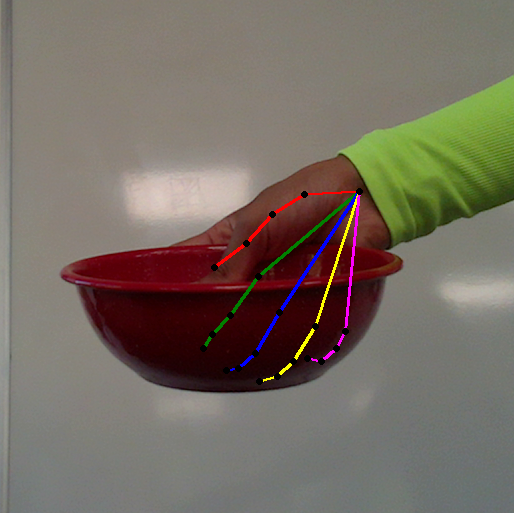}} \\[-0.3cm]
    
    \caption{\textbf{2D hand keypoints results from DOPE.} The left column shows successful examples, typically when there is little occlusion by the object, while the right column shows overall failure cases which occur when the objects largely occlude the hand.}
    \label{fig:dope1}
\end{figure}

\noindent \textbf{Additional Analysis.} 
In Figure~\ref{fig:dope1}, we show qualitative results of the DOPE hand keypoint detector, in both successful (left) and failure cases (right). We can see that in the case of a frame without large occlusion, the hand pose is well estimated. However, when large occlusions occur, the detector wrongly estimates the hand pose, which impacts the camera pose estimation.

\begin{figure}[ht]
    \centering
    \includegraphics[width=.5\textwidth]{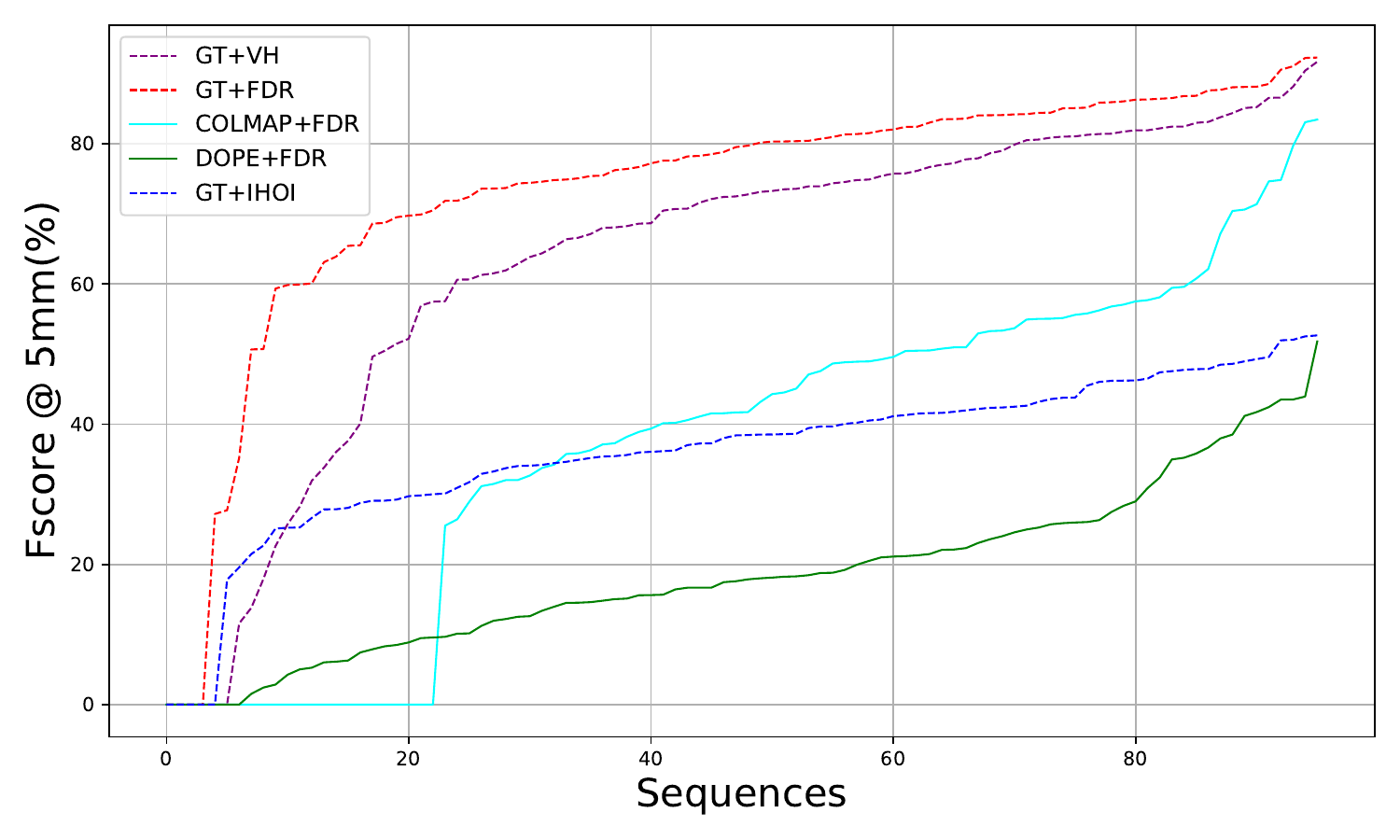}
    \vspace{-0.8cm}
    \caption{\textbf{Fscore@5mm over the dataset sorted by increasing order for various methods.} The closer the curve is to the top-left, the better the method. }
    \label{fig:fscore_5mm}
\end{figure}

Finally, we provide a detailed study of the F-Score over the whole dataset for various setups in~\cref{fig:fscore_5mm}. We can see that both \mvs methods, VH and FDR attain good Fscore using the annotated poses (\gt) while IHOI tends to lack accurate and complete reconstructions with the 5mm threshold. Also, DOPE-based rigid transformation estimations tend to provide worse reconstruction scores but are much more reliable than COLMAP, which fails completely on more sequences.

\begin{table}[]
\resizebox{\linewidth}{!}{
\begin{tabular}{cccc}
\hline 
\multirow{2}{*}{object size} & \multicolumn{3}{c}{Per-Frame Relative Pose Quality (\%) $\uparrow$} \\ 
& @($2\text{cm}\&4^\circ$) & @($5\text{cm}\&10^\circ$) & @($10\text{cm}\&20^\circ$) \\

\midrule
small              & 9.96\std{15.60} & 36.90\std{33.57} & 74.50\std{31.20} \\ 
larger          & 6.58\std{5.17} & 28.10\std{20.85} & 69.50\std{25.21}   \\ 
\hline
\end{tabular}
}
\vspace{-0.2cm}
\caption{\textbf{Rigid transformation evaluation obtained from COLMAP~\cite{schoenberger2016sfm}} with the per-frame relative pose quality, \ie the percentages of frames where the error is below a given threshold, averaged across sequences. 
}
\label{tab:colmaprelpose_size_newmet_perframe}
\end{table}

\begin{table}[]
\resizebox{\linewidth}{!}{
\begin{tabular}{cccc}
\hline 
\multirow{2}{*}{object size} & \multicolumn{3}{c}{Per-Frame Relative Pose Quality (\%) $\uparrow$} \\ 
& @($2\text{cm}\&4^\circ$) & @($5\text{cm}\&10^\circ$) & @($10\text{cm}\&20^\circ$) \\
\midrule
small              & 2.06\std2.13  & 14.20\std9.59 & 40.10\std19.70 \\ 
larger         & 1.70\std1.82 & 9.75\std6.66 & 30.80\std17.50    \\ 
\hline
\end{tabular}
}
\vspace{-0.2cm}
\caption{\textbf{Rigid transformation evaluation obtained from hand joints estimates from DOPE~\cite{weinzaepfel2020dope}}  with the per-frame relative pose quality, \ie the percentages of frames where the error is below a given threshold, averaged across sequences.}
\label{tab:doperelpose_size_newmet_perframe}
\end{table}

\begin{table}[]
\resizebox{\linewidth}{!}{%
\begin{tabular}{ccccc}
\hline
\multirow{2}{*}{method} &  object & acc. ratio & comp. ratio & Fscore \\ 
& size & @5mm (\%) $\uparrow$ & @5mm (\%) $\uparrow$ & @5mm (\%) $\uparrow$ \\
\midrule
\multirow{2}{*}{COLMAP+HHOR}& small              & 32.87                         & 33.57                          & 33.07                     \\ 
&larger         & \textbf{54.35}                         & \textbf{56.14}                          & \textbf{55.00}                     \\  
\midrule
\multirow{2}{*}{DOPE+HHOR}&small              & \textbf{40.87}                         & \textbf{42.48}                          & \textbf{40.96}                     \\ 
& larger          & 37.67                         & 41.03                          & 38.73                     \\ \hline
\end{tabular}%
}
\vspace{-0.2cm}
\caption{ \textbf{Hand-Object Reconstruction evaluation of \sigg~\cite{huang2022hhor} vs object size}.}
\label{tab:hhordoperecon_size}
\end{table}

\begin{table}[]
\resizebox{\linewidth}{!}{%
\begin{tabular}{cccc}
\hline 
\multirow{2}{*}{texture type} & \multicolumn{3}{c}{Per-Frame Relative Pose Quality $\uparrow$} \\ 
& @($2\text{cm}\&4^\circ$) & @($5\text{cm}\&10^\circ$) & @($10\text{cm}\&20^\circ$) \\
\midrule
textured              & 6.69\std4.6  & 29.2\std21.1 & 71.5\std25.9\\ 
less textured          & 9.92\std15.8 & 33.7\std30.9  & 73.1\std28.5    \\ 
\hline
\end{tabular}%
}
\vspace{-0.2cm}
\caption{\textbf{Rigid transformation evaluation obtained from COLMAP~\cite{schoenberger2016sfm}} with the per-frame relative pose quality, \ie the percentages of frames where the error is below a given threshold, averaged across sequences.}
\label{tab:colmaprelpose_tex_newmet}
\end{table}

\begin{table}[]
\resizebox{\linewidth}{!}{%
\begin{tabular}{cccc}
\hline
\multirow{2}{*}{texture type} & acc. ratio & comp. ratio & Fscore \\ 
& @5mm (\%) $\uparrow$ & @5mm (\%) $\uparrow$ & @5mm (\%) $\uparrow$ \\
\midrule
textured              & \textbf{51.87}                         & \textbf{47.64}                          & \textbf{49.47}                     \\ 
less-textured          & 36.19                         & 33.06                          & 34.46                     \\ \hline
\end{tabular}%
}
\vspace{-0.2cm}
\caption{ \textbf{Hand-Object Reconstruction evaluation of COLMAP + HHOR~\cite{huang2022hhor} vs object texture}.}
\label{tab:hhorcolmaprecon_tex}
\end{table}

\subsection{Single-image HO reconstruction}
\label{sub:ihoi}

\begin{figure}
\center
\includegraphics[width=0.48\linewidth]{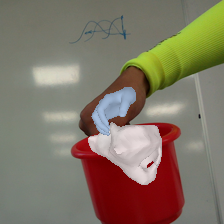}
\includegraphics[width=0.48\linewidth]{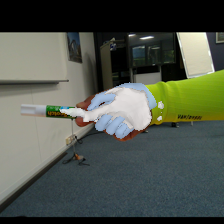}\\
\includegraphics[width=0.48\linewidth]{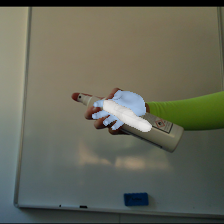}
\includegraphics[width=0.48\linewidth]{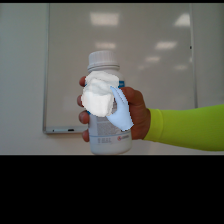} \\[-0.3cm]

\caption{
\textbf{Failure cases of single-image reconstructions} using FrankMocap~\cite{frankmocap} and \cite{ye_whats_2022}.
The method of~\cite{ye_whats_2022} struggles with unusual objects or grasps, such as an object held at the tip of the fingers or large objects.}
\label{fig:ihoi_single_img}
\end{figure}

\begin{figure}
\center
\includegraphics[width=0.24\linewidth]{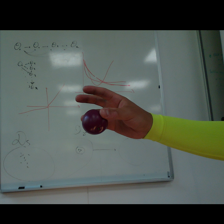}
\includegraphics[width=0.24\linewidth]{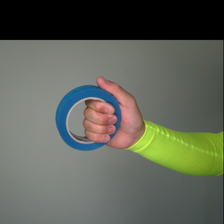}
\includegraphics[width=0.24\linewidth]{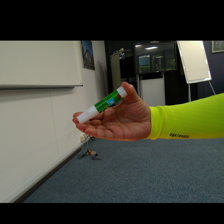}
\includegraphics[width=0.24\linewidth]{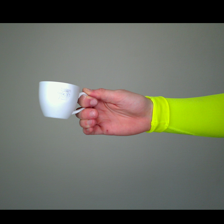} \\
\includegraphics[width=0.24\linewidth]{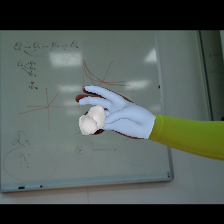}
\includegraphics[width=0.24\linewidth]{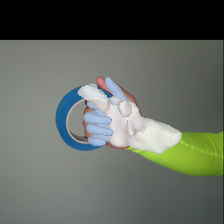}
\includegraphics[width=0.24\linewidth]{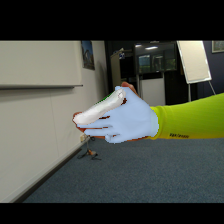}
\includegraphics[width=0.24\linewidth]{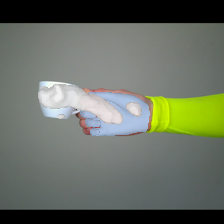} \\
\includegraphics[width=0.24\linewidth]{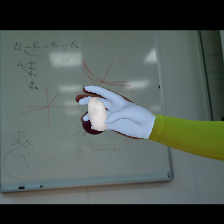}
\includegraphics[width=0.24\linewidth]{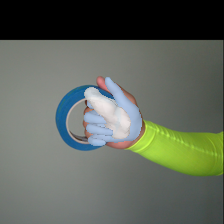}
\includegraphics[width=0.24\linewidth]{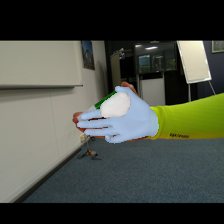}
\includegraphics[width=0.24\linewidth]{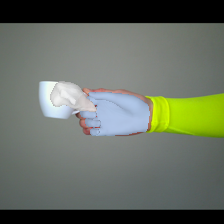} \\[-0.3cm]
\caption{
\textbf{Fusing single-image reconstructions} contributes to smoothing the results, but sometimes degrades the reconstruction. 
\textit{Top:} input image.
\textit{Middle:} result of a single-image reconstruction method~\cite{ye_whats_2022}.
\textit{Bottom:} result from fusing SDFs produced by~\cite{ye_whats_2022}  on several frames of a given sequence. SDFs are estimated in a  canonical hand space (with a fixed wrist position), and fused directly in that space.}
\label{fig:ihoi_sdf}
\end{figure}

We provide in~\cref{fig:ihoi_single_img} some qualitative reconstruction results on a few objects of the dataset. In Section 5.2, we compare the single-view reconstruction method of Ye \etal~\cite{ye_whats_2022} (IHOI) to multiview baselines that take the whole video sequence as input. For a fairer comparison, we slightly modify IHOI for video processing: the method is run independently on multiple, evenly-distributed frames of the sequence. From these single-image reconstructions, we extract the raw object-SDFs (signed distance functions), prior to the meshing step, and average them directly in the canonical reconstruction space, centered around the wrist. Figure~\ref{fig:ihoi_sdf} shows the effect of this temporal aggregation of SDFs on different objects.
In Table~\ref{tab:ihoi_recon}, we show a quantitative comparison of this modified approach, dubbed IHOI+temp, with baseline IHOI, following the evaluation setup presented in Section 5.2 (and Table 3).

\begin{table*}[!h]
	\centering
		\begin{tabular}{l|ccccc}
			\specialrule{.1em}{.05em}{.05em}
			Reconstruction &  Acc. & Comp. & Acc. ratio  & Comp. ratio  & Fscore  \\ 
			Method & (cm) $\downarrow$ &  (cm) $\downarrow$ &  @5mm (\%) $\uparrow$ & @5mm (\%) $\uparrow$ & @5mm (\%) $\uparrow$ \\ 
            \midrule
            IHOI~\cite{ye_whats_2022} & 0.798  & \textbf{1.344} & 41.77 & \textbf{37.80} & \textbf{39.32} \\ 
			IHOI+temp & \textbf{0.757}  & 1.580 & \textbf{42.88} & 34.93 & 38.08 \\ 
            \bottomrule
		\end{tabular}%
\vspace{-0.3cm}
\caption{
\textbf{Comparison of baseline IHOI~\cite{ye_whats_2022} with our temporal extension (IHOI+temp).} On average, combining SDFs from multiple frames improves accuracy but degrades completeness.
}
\label{tab:ihoi_recon}
\vspace{-0.5cm}
\end{table*}

\section{Limitations}
We assume that the hand remains static with respect to the object i.e. the hand pose remains the same with respect to the object throughout the sequence. This assumption enabled to do category-agnostic hand-object reconstruction. However, this limits the reconstruction baselines to dynamic hand-object manipulation scenarios. In summary, this assumption is reasonable in terms of application and an important step towards dynamic object-agnostic HO reconstruction.
{\small
\bibliographystyle{ieee_fullname}

}